\documentclass[accepted]{uai2025} 
                        
\usepackage[american]{babel}

\usepackage{natbib} 
\usepackage{tikz} 

\usepackage{algorithm}
\usepackage[indLines=true]{algpseudocodex}

\usepackage{graphicx}
\usepackage{subfigure}

\usepackage{wrapfig}
\usepackage{multirow}
\usepackage{booktabs} 

\usepackage{amsmath}
\usepackage{mathtools}

\usepackage{amssymb}
\usepackage{bm}
\usepackage{nicefrac}

\usepackage{amsthm}
\usepackage{thmtools}

\theoremstyle{plain}
\newtheorem{theorem}{Theorem}[section]

\theoremstyle{definition}
\newtheorem{definition}[theorem]{Definition}

\theoremstyle{remark}

\newtheorem{prop}{Proposition}

\usepackage{etoolbox}
\usepackage{xparse}

\allowdisplaybreaks    

\numberwithin{equation}{section}  

\usepackage{amsmath,amsfonts,bm}

\def\eqref#1{equation~\ref{#1}}

\def\1{\bm{1}}

\def\vm{{\bm{m}}}

\def\vw{{\bm{w}}}
\def\vx{{\bm{x}}}

\def\mS{{\bm{S}}}

\def\mSigma{{\bm{\Sigma}}}

\DeclareMathAlphabet{\mathsfit}{\encodingdefault}{\sfdefault}{m}{sl}
\SetMathAlphabet{\mathsfit}{bold}{\encodingdefault}{\sfdefault}{bx}{n}

\newcommand{\Ltwo}[2]{L^2(#1, #2)}
\newcommand{\innerProd}[2]{\langle #1, #2 \rangle}

\usepackage{amsmath,amsfonts,amssymb,bm,mathtools}
\usepackage{xcolor}

\newcommand{\measureP}{\mathbb{P}}
\newcommand{\measureQ}{\mathbb{Q}}

\newcommand{\eye}{\mathbb{I}}

\newcommand{\br}[1]{\mathopen{}\left(#1\right)\mathclose{}} 
\newcommand{\sqbr}[1]{\mathopen{}\left[#1\right]\mathclose{}} 

\newcommand{\set}[1]{\left\{#1\right\}}

\newcommand{\norm}[1]{\left\lVert#1\right\rVert} 

\newcommand{\expect}[2]{{\mathop{{}\mathbb{E}}}_{#2}\sqbr{#1}} 

\newcommand{\grad}[2]{\nabla_{#1}#2}
\newcommand{\tp}{^\top} 

\newcommand{\prob}[1]{p\br{#1}} 
\newcommand{\g}{\,|\,} 
\newcommand{\kl}[2]{D_{\mathrm{KL}}\br{#1\,\|\,#2}} 
\newcommand{\regkl}[2]{D^{\gamma}_{\mathrm{KL}}\br{#1\,\|\,#2}} 
\newcommand{\regklhat}[2]{\hat{D}^{\gamma}_{\mathrm{KL}}\br{#1\,\|\,#2}} 

\newcommand{\real}{\mathbb{R}}

\newcommand{\gaussian}[2]{\mathcal{N}\br{#1,#2}}
\newcommand{\gaussianx}[3]{\mathcal{N}\br{#1\g #2,#3}}

\newcommand{\ie}{i.e.\ }

\makeatletter
\newcommand{\oset}[3][1ex]{%
  \mathrel{\mathop{#3}\limits^{
    \vbox to#1{\kern-2\ex@
    \hbox{$\scriptstyle#2$}\vss}}}}
\makeatother

\usepackage{blindtext}

\usepackage[
  capitalize,
  noabbrev,
]{cleveref}

\title{Well-Defined Function-Space Variational Inference\\in Bayesian Neural Networks via Regularized KL-Divergence}
\author[1]{\href{mailto:<tristan.cinquin@uni-tuebingen.de>?Subject=Your UAI 2025 paper}{Tristan Cinquin}{}}
\author[1]{Robert Bamler}
\affil[1]{%
    University of Tübingen, Germany
}  

\begin{document}
\maketitle
\begin{abstract}
Bayesian neural networks (BNN) promise to combine the predictive performance of neural networks with principled uncertainty modeling crucial for safety-critical systems and decision making.
However, posterior uncertainties depend on the choice of prior, and finding informative priors in weight-space has proven difficult.
This has motivated variational inference (VI) methods that pose priors directly on the function represented by the BNN rather than on weights.
In this paper, we address a fundamental issue with such function-space VI approaches pointed out by \citet{burt2020understanding}, who showed that the objective function (ELBO) is negative infinite for most priors of interest.
Our solution builds on \emph{generalized} VI with the regularized KL divergence and is, to the best of our knowledge, the first well-defined variational objective for inference in BNNs with Gaussian process (GP) priors.
Experiments show that our method successfully incorporates the properties specified by the GP prior, and that it provides competitive uncertainty estimates for regression, classification and out-of-distribution detection compared to BNN baselines with both function and weight-space priors.
\end{abstract}

\section{Introduction}
\begin{figure*}[t]
    \centering
    \resizebox{\linewidth}{!}{
    \includegraphics[width=\linewidth]{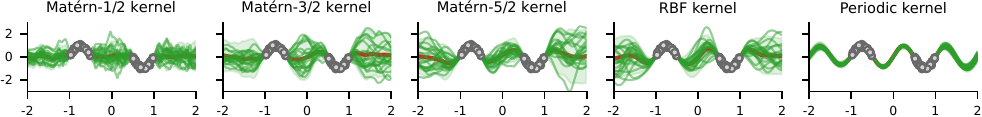}
    }
    \caption{
    Inference with our GFSVI on synthetic data (gray circles) with Gaussian process priors encoding different properties such as smoothness (increasing from Matérn-1/2 to RBF) and periodicity (last panel). 
    }
    \label{fig:fsvi_prior_varying_smoothness}
\end{figure*}
%
Neural networks have shown impressive results in many fields but fail to provide well-calibrated uncertainty estimates, which are essential in applications associated with risk, such as healthcare \citep{medicineBDL} or finance \citep{analystRecommendationBML}.
Bayesian neural networks (BNNs) offer to combine the scalability and predictive performance of neural networks with principled uncertainty modeling by explicitly capturing epistemic uncertainty, which results from finite training data.
While the choice of prior strongly affects posterior uncertainties, specifying informative priors on BNN weights has proven difficult and is hypothesized to have limited their practical applicability \citep{knoblauch2019generalized,tran2022NeedGoodfuncpriorforBNNs}.
For instance, the default isotropic Gaussian prior, which is often chosen for tractability rather than for the beliefs it carries \citep{knoblauch2019generalized}, is known to have pathological behavior in some cases \citep{cinquin2021pathologies,tran2022NeedGoodfuncpriorforBNNs}.
A promising approach to solve this issue is to place priors directly on the function represented by the BNN instead of the weights.
Function-space priors allow incorporating interpretable knowledge, for instance using the Gaussian Process (GP) literature to improve prior design and selection \citep{williams2006gaussian}. 

A recent line of work has focused on using function-space priors in BNNs with variational inference (VI) \citep{sun2018functional}. 
VI is appealing because of its successful application to BNNs, its flexibility in terms of approximate posterior parameterization, and its scalability to large datasets and models \citep{hoffman13SVI,blundell2015weight}.    
Unfortunately, for BNNs with function-space priors, the Kullbach-Leibler (KL) divergence term in the VI objective (ELBO) involves two intractabilities: (i)~a supremum over infinitely many subsets and (ii)~access to the density of the distribution of the BNN's function, which has no closed-form expression. 
\citet{sun2018functional} propose to address problem~(i) by approximating the supremum in the KL divergence by an expectation, and problem~(ii) by using implicit score function estimators (which make this method difficult to use in practice \citep{ma2021funcVIspg}).  
However, the problem is actually more severe.
Not only is the KL divergence intractable, it is infinite in most cases of interest \citep{burt2020understanding}, such as when the prior is a non-degenerate GP or a BNN with a different architecture.
Thus, in these (and many more) situations, the KL divergence cannot even be approximated.
As a consequence, more recent work abandons using BNNs and instead uses deterministic neural networks to parameterize basis functions \citep{ma2021funcVIspg} or a GP mean \citep{wild2022gvi}.
The only prior work \citep{rudner2022fsvi} that overcomes the issue pointed out by \citet{burt2020understanding} does so by deliberately limiting itself to cases where the KL divergence is known to be finite (by defining the prior as the pushforward of a weight-space distribution).
Therefore, the method by \citet{rudner2022fsvi} suffers from the same issues regarding prior specification as other weight-space inference method.

In this paper, we address the argument by \citet{burt2020understanding} that VI does not provide a valid objective for inference in BNNs with genuine function-space priors, and we propose to apply the framework of generalized VI \citep{knoblauch2019generalized}.
We present a simple method for function-space inference with GP priors that builds on the regularized KL divergence~\citep{quang2019regularizedKL}, which generalizes the conventional KL divergence and is finite for any pair of Gaussian measures.
We obtain a Gaussian measure for the variational posterior by considering the linearized BNN from \citet{rudner2022fsvi}, and we are free to choose a function-space prior from a large set of GPs which have an associated Gaussian measure on the considered function space. 
While the regularized KL divergence is still intractable, it can be consistently estimated from samples with a known error bound.
We find that our method effectively incorporates the beliefs specified by GP priors (see \cref{fig:fsvi_prior_varying_smoothness}, discussed further in \cref{sec:experiments}) and that it yields competitive performance compared to BNN baselines.
To the best of our knowledge, our method is the first to provide a well-defined objective for function-space inference in BNNs with informative GP priors. 
Our contributions are summarized below:
\begin{enumerate}
    \item We use generalized VI with the \emph{regularized} KL divergence to mitigate the issue of an infinite KL divergence when using VI in BNNs with function-space priors.
    \item We present a new and well-defined objective for function-space inference in the linearized BNN with GP priors, resulting in a simple algorithm.
    \item We show that our method accurately captures structural properties specified by the GP prior and provides competitive uncertainty estimates for regression, classification, and out-of-distribution detection compared to baselines with both function- and weight-space priors.
\end{enumerate}
The paper is structured as follows: \cref{sec:background} introduces function-space VI and the regularized KL divergence; \cref{sec:methods} presents our method for generalized function-space VI (GFSVI) in BNNs; \cref{sec:experiments} reports experimental results; \cref{sec:related_work} discusses related work; and \cref{sec:discussion} concludes.
%
%
\section{Background}
\label{sec:background}
In this section, we provide background on function-space variational inference in BNNs and discuss the fundamental issue of an infinite KL divergence.
We then introduce the regularized KL divergence, which is the basis for our solution presented in Section~\ref{sec:methods}. 
\subsection{Function-space VI in BNNs}
\label{sec:fsvi_in_bnns}
%
We consider a neural network $f({\,\cdot\,; \vw})$ with weights $\vw \in \real^p$, and a data set $\mathcal{D} = \{(\vx_i, y_i)\}_{i=1}^N$ with features~$\vx_i \in \mathcal{X} \subset \real^d$ and values~$y_i \in \mathcal{Y}$.
Bayesian Neural Networks are specified further by a likelihood function $\prob{\mathcal D \g \vw} = \prod_{i=1}^N \prob{y_i \g f(\vx_i;\vw)}$ and---traditionally---a prior $\prob{\vw}$ on the weights, and one seeks the posterior distribution $\prob{\vw \g \mathcal{D}} \propto \prob{\mathcal D \g \vw} \, \prob{\vw}$.
The method proposed in this paper builds on variational inference, which approximates $\prob{\vw \g \mathcal D}$ with a variational distribution $q_\phi\br{\vw}$, whose variational parameters~$\phi$ maximize the evidence lower bound (ELBO),
\begin{align}\label{eq:elbo-weight space}
    \mathcal L(\phi) := \expect{\log \prob{\mathcal D \g \vw}}{q_\phi\br{\vw}} - \kl{q_\phi}{p}
\end{align}
where $D_\text{KL}$ is the Kullback-Leibler divergence,
\begin{equation}\label{eq:kl-weight space}
    \kl{q_\phi}{p} := \expect{\log\br{q_\phi\br{\vw} \big/ \prob{\vw}}}{q_\phi\br{\vw}}.
\end{equation}
At test time, we approximate the predictive distribution for given features~$\vx^*$ as $\prob{y^* \g \vx^*} = \expect{\big.\prob{y^* \g f(\vx^*;\vw)}}{\prob{\vw \g \mathcal{D}}} \approx \expect{\big.\prob{y^* \g f(\vx^*;\vw)}}{q_\phi(\vw)}$.
\paragraph{Function-space variational inference.}
%
Since neural network weights are not interpretable, we replace the weight-space prior $p(\vw)$ with a prior~$\measureP$ directly on the function $f\br{\,\cdot\,;\vw}$, which we denote simply as~$f$ when there is no ambiguity.
Here, the symbol~$\measureP$ denotes a probability measure that does not admit a density since the function space is infinite-dimensional.
We compute the expected log-likelihood as in the first term of Eq.~\ref{eq:elbo-weight space}.
For the KL-term (Eq.~\ref{eq:kl-weight space}), a naive VI-method would use the pushforward of $q_\phi(\vw)$ along $\vw\mapsto f\br{\,\cdot\,;\vw}$, which defines the variational measure~$\measureQ_\phi$, resulting in the ELBO in function space,
\begin{equation}\label{eq:elbo-function space}
    \mathcal L(\phi) := \expect{\log \prob{\mathcal D \g \vw}}{q_\phi\br{\vw}} - \kl{\measureQ_\phi}{\measureP}
\end{equation}
with $D_\text{KL}$ the KL divergence between measures
\begin{equation}\label{eq:kl-function space}
    \kl{\measureQ_\phi}{\measureP} := \int\! \log \br{\!\frac{d\measureQ_\phi}{d\measureP}(f)\!} d\measureQ_\phi.
\end{equation}
Here, the Raydon-Nikodym derivative $d\measureQ_\phi / d\measureP$ generalizes the density ratio from Eq.~\ref{eq:kl-weight space}.
Like Eq.~\ref{eq:elbo-weight space}, the ELBO in Eq.~\ref{eq:elbo-function space} is a lower bound on the evidence \citep{burt2020understanding}.
In fact, if $\measureP$ is the push-forward of $p(\vw)$ then Eq.~\ref{eq:elbo-function space} is a tighter bound than Eq.~\ref{eq:elbo-weight space} by the data processing inequality, $\kl{\measureQ_\phi}{\measureP} \leq \kl{q_\phi}{p}$.
However, we motivated function-space VI to avoid weight-space priors, and in this case the bound in Eq.~\ref{eq:elbo-function space} can be looser.
We will indeed see below that the bound becomes infinitely loose in practice, and we thus propose a different objective in Section~\ref{sec:methods}.

Two intractabilities prevent directly maximizing the ELBO in function space (Eq~\ref{eq:elbo-function space}).
First, it is not obvious how to evaluate or estimate the KL divergence between two measures in Eq~\ref{eq:kl-function space}.
\citet{sun2018functional} showed that it can be expressed as a supremum of KL divergences between finite-dimensional distributions,
\begin{equation} \label{eq:sun_kl}
    \kl{\measureQ_\phi}{\measureP} = \sup_{\vx \in \mathcal{X}^M, M \in \mathbb{N}} \kl{q_\phi(f(\vx))}{p(f(\vx))}.
\end{equation}
Here, $\vx = \{\vx^{(i)}\}_{i=1}^M \in \mathcal{X}^M$ is a set of $M$ points in feature space~$\mathcal X$, and $q_\phi(f(\vx))$ and $p(f(\vx))$ are densities of the marginals of $\measureQ_\phi$ and~$\measureP$ on~$\{f(\vx^{(i)})\}_{i=1}^M$ respectively.
\citet{sun2018functional} approximates the supremum over infinitely many sets by an expectation, and \citet{rudner2022fsvi} estimates it from samples.

Second, we cannot express the pushforward measure~$\measureQ_\phi$ in closed form because the neural network is nonlinear. 
Previous work has proposed to mitigate this issue using implicit score function estimators \citep{sun2018functional} or a linearized BNN~$f_L$ to obtain a closed-form Gaussian variational measure \citep{rudner2022ContinualLearningFSVI, rudner2022fsvi}.
Our proposal in Section~\ref{sec:methods} follows the linearized BNN approach as it only minimally modifies the BNN, preserving most of its inductive bias \citep{maddox2021FastAdapt} while considerably simplifying the problem by turning the pushforward of $q_\phi(\vw)$ into a GP.
More specifically, we consider a Gaussian variational distribution $q_{\phi}\br{\vw} = \gaussian{\vm}{\mS}$ with parameters $\phi = \set{\vm, \mS}$, 
and we define a linearized BNN~$f_L$ by linearizing~$f$ as a function of the weights around~$\vw=\vm$,
\begin{equation}\label{eq:linearized_nn}
    f_L(\vx; \vw) := f(\vx; \vm) + J(\vx; \vm)(\vw - \vm)
\end{equation}
with $J(\vx; \vm) = \grad{\!\vw}{f({\vx; \vw})}|_{\vw=\vm}$. Thus, $\vw\sim q_\phi(\vw)$ implies $f_L(\vx;\vw) \sim \gaussian{f(\vx; \vm)}{J(\vx;\vm) \mS J(\vx;\vm)\tp}$ for all~$\vx \in \mathcal{X}$, and so the function $f_L(\,\cdot\,; \vw)$ is a degenerate GP (as $\operatorname{rank}(J(\,\cdot\,;\vm) \mS J(\,\cdot\,;\vm)\tp) \leq p <\infty$),
\begin{equation}\label{eq:linearized-q}
    f_L \sim \mathcal{GP}\br{f(\,\cdot\,; \vm), J(\,\cdot\,;\vm) \mS J(\,\cdot\,;\vm)\tp}.
\end{equation}
\paragraph{$\kl{\measureQ_\phi}{\measureP}$ is infinite in most relevant cases.}
\citet{burt2020understanding} point out an even more severe issue of function-space VI in BNNs: $\kl{\measureQ_\phi}{\measureP}$ (Eq.~\ref{eq:kl-function space}) is in fact infinite in most relevant cases, in particular for non-degenerate GP-priors.
Thus, approximating $\kl{\measureQ_\phi}{\measureP}$ in these settings is futile.
Their proof is somewhat involved, but the fundamental reason for $\kl{\measureQ_\phi}{\measureP}=\infty$ is that $\measureQ_\phi$~has support on a finite-dimensional submanifold of the infinite-dimensional function space, while the measure~$\measureP$ induced by a (non-degenerate) GP prior has support on the entire function space.
That such a dimensionality mismatch can lead to infinite KL divergence can already be seen in a finite-dimensional example: consider the KL-divergence between two Gaussians in~$\mathbb R^n$ for~$n\geq 2$, one of which has support on the entire~$\mathbb R^n$ (i.e., its covariance matrix~$\mSigma_1$ has full rank) while the other one has support only on a proper subspace of~$\mathbb R^n$ (i.e., its covariance matrix~$\mSigma_2$ is singular).
The KL divergence between multivariate Gaussians has a closed form expression (Eq.~\ref{eq:reg-kl-estimator} with ${\gamma=0}$) that contains $\log\det\br{\mSigma_2^{-1}\mSigma_1}$, which is infinite for singular~$\mSigma_2$.

We find that the fact that $\kl{\measureQ_\phi}{\measureP}=\infty$ has severe practical consequences even when the KL divergence is only estimated from finite samples.
It naturally explains the stability issues discussed in Appendix~D.1 of \citet{sun2018functional}
(we compare the authors' solution to this stability issue to our method in \Cref{sec:gfsvi-comparison}).
Surprisingly, similar complications arise even in the setup by \citet{rudner2022fsvi}, which performs VI in function space with the pushforward of a weight-space prior.
While this makes the KL divergence technically finite because prior and variational posterior have the same support, numerical errors lead to mismatching supports and thus to stability issues even there.

In summary, the ELBO for VI in BNNs is not well-defined for most interesting function-space priors.
In \cref{sec:methods}, we propose a solution by using the so-called regularized KL divergence, which we introduce next.

\subsection{Regularized KL divergence}
\label{sec:reg_kl}
Our solution to the negative infinite function-space ELBO builds on a regularized KL divergence, which is expressed in terms of Gaussian measures for the variational posterior and prior.
We obtain these Gaussian measures from GPs.
We first discuss under which conditions a GP induces a Gaussian measure, and then present the regularized KL divergence.
\paragraph{Gaussian measures and Gaussian processes.}
\label{par:gaussian-measures}
The regularized KL divergence is defined in terms of Gaussian measures, and thus we need to verify that the GP variational posterior induced by the linearized BNN (Eq.~\ref{eq:linearized-q}) has an associated Gaussian measure.
We consider the Hilbert space $\Ltwo{\mathcal{X}}{\rho}$ of square-integrable functions with respect to a probability measure $\rho$ on a compact set $\mathcal{X} \subset \real^d$, with inner product $\innerProd{f}{g} = \int_\mathcal{X} f(x) g(x) d\rho(x)$. 
This assumption is not restrictive since we can typically bound the region in feature space that contains the data and any points where we might want to evaluate the BNN. 
\begin{definition}[Gaussian measure, \citet{kerrigan2023diffusion}, Definition 1]\label{def:gm}
Let $(\Omega, \mathcal{B}, \measureP)$ be a probability space. A measurable function $F: \Omega \mapsto \Ltwo{\mathcal{X}}{\rho}$ is called a Gaussian random element (GRE) if for any $g \in \Ltwo{\mathcal{X}}{\rho}$ the random variable $\innerProd{g}{F}$ has a Gaussian distribution on $\real$. 
For every GRE $F$, there exists a unique mean element $m \in \Ltwo{\mathcal{X}}{\rho}$ and a finite trace linear covariance operator $C: \Ltwo{\mathcal{X}}{\rho} \mapsto \Ltwo{\mathcal{X}}{\rho}$ such that $\innerProd{g}{F} \sim \gaussian{\innerProd{g}{m}}{\innerProd{Cg}{g}}$ for all $g \in \Ltwo{\mathcal{X}}{\rho}$.
The pushforward of $\measureP$ along~$F$, denoted $\measureP^F := F_{\#} \measureP$, is a Gaussian measure on $\Ltwo{\mathcal{X}}{\rho}$.
\end{definition}

Gaussian measures generalize Gaussian distributions to infinite-dimensional function spaces where measures do not have associated densities since there is no Lebesgue measure. 
Following \citet{wild2022gvi}, we notate the Gaussian measure obtained from the GRE $F$ with mean element $m$ and covariance operator $C$ as $\measureP^F := \gaussian{m}{C}$.
GPs provide a practical tool to specify Gaussian measures via mean and covariance functions \citep{kerrigan2023diffusion}.
A GP $f \sim \mathcal{GP}(\mu, K)$ has an associated Gaussian measures in $\Ltwo{\mathcal{X}}{\rho}$ if its mean function satisfies $\mu \in \Ltwo{\mathcal{X}}{\rho}$ and its covariance function $K$ is trace-class, i.e., if $\int_{\mathcal{X}} K(x, x) d \rho \br{x} < \infty$ \citep[Theorem~1]{wild2022gvi}.
The GP variational posterior induced by the linearized BNN satisfies both properties as neural networks are well-behaved functions on the compact $\mathcal{X} \subset \real^d$.
It thus induces a Gaussian measure $\measureQ_\phi^F \sim \gaussian{m_Q}{C_Q}$ with mean element $m_Q = f(\,\cdot\,; \vm)$ and covariance operator $C_Q g \br{\cdot} = \int_{\mathcal{X}} J(\,\cdot\,; \vm) \mS J(\vx', \vm)\tp g(\vx') d\rho\br{\vx'}$.
The infinite KL divergence discussed in Section~\ref{sec:fsvi_in_bnns} is easier to prove for the special case of Gaussian measures, and we provide the proof in \cref{sec:infinite_kl}.
\begin{definition}[Regularized KL divergence, \citet{quang2022gpkl} Definition 5]
Let $\nu_1=\gaussian{m_1}{C_1}$ and $\nu_2=\gaussian{m_2}{C_2}$ be two Gaussian measures with $m_1, m_2 \in \Ltwo{\mathcal{X}}{\rho}$ and $C_1, C_2$ bounded, self-adjoint, positive and trace-class linear operators on $\Ltwo{\mathcal{X}}{\rho}$. 
Let $\gamma \in \real_{>0}$ be fixed. 
The regularized KL divergence is defined as follows,
\begin{align}
&\regkl{\nu_1}{\nu_2} := \frac{1}{2} \innerProd{m_1 - m_2}{(C_2 + \gamma \eye)^{-1} \br{m_1 - m_2}} \nonumber\\
&\qquad+ \frac{1}{2} \operatorname{Tr}_X \left[ \br{C_2 + \gamma \eye}^{-1} \br{C_1 + \gamma \eye} - \eye \right] \nonumber\\
&\qquad- \frac{1}{2} \log \operatorname{det}_X \left[ \br{C_2 + \gamma \eye}^{-1} \br{C_1 + \gamma \eye}\right]. \label{eq:regkl-definition}
\end{align}
\end{definition}
Here $\operatorname{Tr}_X$ and $\operatorname{det}_X$ are the extended trace and extended Fredholm determinant \citep{quang2022gpkl}.
For any $\gamma > 0$, the regularized KL divergence is well-defined and finite (following \citet[Proposition 1]{quang2017infinitedimensionallogdeterminantdivergencesii}), even if the Gaussian measures are singular \citep{quang2019regularizedKL}.
It converges to the conventional KL divergence (if it is well-defined) for $\gamma \to 0$ (\citealp[Theorem 6]{quang2022gpkl}). 
Furthermore, if the Gaussian measures $\nu_1$ and $\nu_2$ are induced by GPs
$\mathcal{GP}(\mu_i, K_i)$ for $i=1,2$, respectively, then $\regkl{\nu_1}{\nu_2}$ is consistently estimated~\citep{quang2022gpkl} by
\begin{align}
    \regklhat{\nu_1}{\nu_2} :=&\, \frac{1}{2} \br{\vm_1 - \vm_2}\tp (\mSigma^{(\gamma)}_2)^{-1} \br{\vm_1 - \vm_2} \nonumber\\
    +&\, \frac{1}{2} \operatorname{Tr}\big[(\mSigma_2^{(\gamma)})^{-1}\mSigma^{(\gamma)}_1 - \eye_M\big] \nonumber\\
    -&\, \frac{1}{2} \log \det \big[(\mSigma_2^{(\gamma)})^{-1}\mSigma^{(\gamma)}_1\big] \label{eq:reg-kl-estimator}
\end{align}
with $\vm_i:=\mu_i(\vx)$ and $\mSigma_i^{(\gamma)}:=K_i(\vx,\vx) + \gamma M \,\eye_M$ where $\mu_i(\vx)$ and $K_i(\vx,\vx)$ are the mean vector and the covariance matrix obtained by evaluating $\mu_i$ and $K_i$ respectively, at measurement points $\vx = \{\vx^{(i)}\}_{i=1}^M \oset[.40ex]{\textup{\tiny i.i.d}}{\sim} \rho(\vx)$. 
The right-hand side of Eq.~\ref{eq:reg-kl-estimator} is the expression for the KL-divergence between Gaussian distributions $\mathcal N({\vm_1},{\mSigma_1^{(\gamma)}})$ and $\mathcal N({\vm_2},{\mSigma_2^{(\gamma)}})$.
\citet{quang2022gpkl} shows that the absolute error of the estimator is bounded by $\mathcal{O}(\sqrt{1/M})$ with high probability with constants depending on $\gamma$ and properties of the GP mean and covariance functions (see \cref{sec:app_reg_kl} for the exact bound).
\section{Generalized function-space VI with the regularized KL divergence}
\label{sec:methods}
This section presents our proposed generalized function-space variational inference (GFSVI) method, which addresses the problem of the infinite KL divergence discussed in \cref{sec:fsvi_in_bnns}, which we take for an indication that VI is too restrictive if one wants to use genuine function-space priors.
We instead consider generalized variational inference \citep{knoblauch2019generalized}, which reinterprets the ELBO in Eq.~\ref{eq:elbo-weight space} as a regularized expected log-likelihood and explores alternative divergences for the regularizer.
Specifically, we propose to use the regularized KL divergence.
This section builds heavily on tools introduced in \cref{sec:background}, which turn out to fit together perfectly: the pushforward of a Gaussian variational distribution in weight-space through the linearized neural network (Eq.~\ref{eq:linearized_nn}) induces a GP variational posterior (Eq.~\ref{eq:linearized-q}) that admits a Gaussian measure on $\Ltwo{\mathcal{X}}{\rho}$. 
Further, selecting a GP prior which has an associated Gaussian measure on $\Ltwo{\mathcal{X}}{\rho}$ allows us to use the regularized KL divergence (Eq.~\ref{eq:regkl-definition}). 
We present GFSVI in \cref{sec:gfsvi} and compare it to prior work in \cref{sec:gfsvi-comparison}.
\subsection{Generalized function-space VI}
\label{sec:gfsvi}
We present a well-defined objective for function-space inference, and a simple algorithm for its optimization.
\paragraph{Objective function.}
We start from the ELBO in Eq.~\ref{eq:elbo-function space}, where we use the Gaussian variational measure~$\measureQ_\phi^F$ induced by the pushforward of a Gaussian variational distribution $q_\phi(\vw) = \gaussianx{\vw}{\vm}{\mS}$ along the linearized network~$f_L$ (Eq.~\ref{eq:linearized_nn}). 
The function-space prior may be any GP that has an associated Gaussian measure~$\measureP^F$ on $\Ltwo{\mathcal{X}}{\rho}$.
We now replace the KL divergence in the ELBO with the regularized KL divergence $D_\text{KL}^\gamma$ (Eq.~\ref{eq:regkl-definition}), which is well-defined and finite for any pair of Gaussian measures.
For a likelihood function $\prob{\mathcal{D} \g \vw} = \prod_{i=1}^N \prob{y_i \g f_L(\vx_i; \vw)}$, we obtain
\begin{equation}\label{eq:objective}
    \mathcal{L}(\phi) \!:=\!\! \sum_{i=1}^N \expect{\log \prob{y_i | f_L(\vx_i; \vw)}}{q_\phi(\vw)} \!-\! \regkl{\measureQ_\phi^F}{\measureP^F}.
\end{equation}
\paragraph{Estimation and optimization.}
The expected log-likelihood (first term in Eq.~\ref{eq:objective}) can be estimated by sampling from $q_\phi(\vw)$.
For a Gaussian likelihood, it can also be computed in closed form as (unlike \citet{rudner2022fsvi}) we use the linearized network~$f_L$
, which made training more stable in our experiments.
We estimate the regularized KL divergence (second term in Eq.~\ref{eq:objective}) using its consistent estimator (see Eq.~\ref{eq:reg-kl-estimator}), with
$\vm_1 = f(\vx; \vm)$,
$\mSigma_1^{(\gamma)} = J(\vx; \vm) \mS J(\vx; \vm)\tp+\gamma M \eye_M$,
$\vm_2 = \mu(\vx)$, and
$\mSigma_2^{(\gamma)} = K(\vx, \vx)+\gamma M \eye_M$,
where $\mu$ and~$K$ are the mean and covariance functions of the GP prior, and $\vx = \{\vx^{(i)}\}_{i=1}^M \oset[.40ex]{\textup{\tiny i.i.d}}{\sim} \rho(\vx)$ are measurement points.
We maximize the estimated objective over the mean~$\vm$ and covariance~$\mS$ of the Gaussian variational distribution $q_\phi(\vw)$, and over any likelihood parameter (e.g., the variance of a Gaussian likelihood), see Algorithm~\ref{alg:fsvi}.
\cref{app:sec_gfsvi_estimator} provides expressions for the estimator with Gaussian and Categorical likelihoods as well as an analysis of their computational complexity.
\paragraph{Technical details ($\gamma$ and $\rho$).}
%
It turns out that increasing~$\gamma$ reduces the influence of the prior on inference (see \cref{fig:regkl_vs_gamma}).
At the same time, $\gamma$~acts as jitter that prevents numerical errors (see \cref{sec:gfsvi-comparison}).
We recommend setting~$\gamma$ large enough to avoid numerical errors but sufficiently small to strongly regularize the objective in Eq.~\ref{eq:objective} (see \cref{fig:fsvi_influence_gamma} in appendix) and setting $M$ to the largest value allowed by the computational budget.
We found that the estimator $\hat D_\text{KL}^\gamma \big(\measureQ_\phi^F \,\big|\!\big|\, {\measureP^F}\big)$ converges quickly to a finite value (especially for smooth kernels, see \cref{fig:regkl_vs_gamma} in appendix), and that GFSVI is robust to a wide range of values (we fixed $\gamma=10^{-10}$).
The probability measure~$\rho$ for $\Ltwo{\mathcal{X}}{\rho}$ has to assign non-zero probability to any open set of $\mathcal{X}$ to regularize the BNN on all of its support.
Following \citet{rudner2022fsvi}, we draw measurement points from a uniform distribution over~$\mathcal{X}$ when using tabular data and explore different configurations (samples from other data sets) for high-dimensional image data (see \cref{app:sec_classification_details}).
\begin{figure*}[t]
    \centering
    \resizebox{\linewidth}{!}{
    \includegraphics[width=\linewidth]{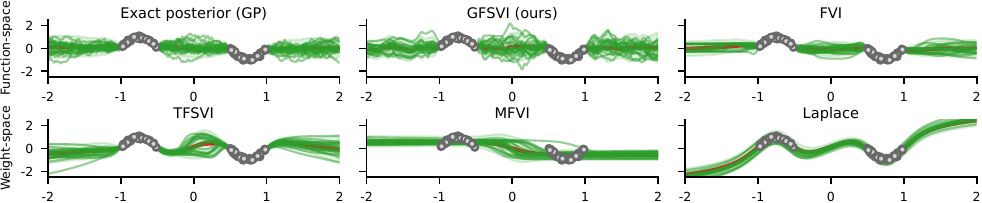}
    }
    \caption{Inference on synthetic data (gray circles) using a Matérn-1/2 prior for function-space methods GFSVI and FVI.
    The proposed GFSVI provides the best approximation of the exact GP posterior.}
    \label{fig:fsvi_matern_vs_baselines}
\end{figure*}
\begin{algorithm}[t]
    \caption{Generalized function-space variational inference (GFSVI)}
    \label{alg:fsvi}
    \begin{algorithmic}[1]
    \Require Linearized BNN $f_L$ with measure $\measureQ_\phi^F$, GP prior $\mathcal{GP}(\mu, K)$ with measure $\measureP^F$, measurement point distribution~$\rho(\vx)$, data $\mathcal{D} = \{(\vx_i, y_i)\}_{i=1}^N$, batch size~$B$, $\gamma > 0$.
    \ForAll{\textnormal{minibatch }$(\vx_\mathcal{B}, y_\mathcal{B}) \sim \mathcal{D}$}
        \State \hbox{Calculate $\hat\ell_1 = \frac{N}{B}\expect{\log \prob{y_\mathcal{B} \g f_L(\vx_\mathcal{B}, \vw)}}{q_{\phi}(\vw)}$;}
        \State \hbox{Draw measurement points $\vx = \{\vx^{(i)}\}_{i=1}^M \stackrel{\!\!\text{i.i.d.}\!\!}{\sim} \rho(\vx)$;}
        \State \hbox{Calculate $\hat\ell_2 = \hat D_\text{KL}^\gamma \big(\measureQ_\phi^F \,\big|\!\big|\, {\measureP^F}\big)$ using $\vx$ (Eq.~\ref{eq:reg-kl-estimator});}
        \State Calculate $\hat{\mathcal{L}}(\phi) = \hat\ell_1 - \hat\ell_2$\;
        \State Update $\phi$ using a step in the direction $\nabla_{\!\phi} \hat{\mathcal{L}}(\phi)$\;
    \EndFor
\end{algorithmic}
\end{algorithm}
\subsection{Connections to prior work}
\label{sec:gfsvi-comparison}
%
TFSVI \citep{rudner2022fsvi} and FVI \citep{sun2018functional} solve stability issues by introducing jitter/white noise, which has a similar effect as the regularization in Eq.~\ref{eq:regkl-definition}.
However, 
TFSVI introduces jitter only to overcome numerical issues and is fundamentally restricted to prior specification in weight space since its function-space prior is the pushforward of a weight-space prior.
Conversely, FVI adds white noise to prevent the KL divergence (Eq.~\ref{eq:kl-function space}) to blow up as $M$ increases. 
However, FVI does not linearize the BNN, and hence does not have access to an explicit variational measure in function space.
This severely complicates the estimation of (gradients of) the KL divergence in FVI, and the authors resort to implicit score function estimators, which make their method difficult to use in practice \citep{ma2021funcVIspg}.
Our proposed GFSVI does not suffer from these difficulties as the variational posterior is an explicit Gaussian measure.
This allows us to estimate the regularized KL divergence without sampling any noise or using implicit score function estimators.
\section{Experiments}
\label{sec:experiments}
In this section, we evaluate our generalized function-space variational inference (GFSVI) method qualitatively on synthetic data and quantitatively on real-world data.
GFSVI accurately captures structural properties specified by the GP prior, and that it performs competitively on regression, classification and out-of-distribution detection tasks.
We also discuss the influence of the BNN's inductive biases.
\paragraph{Baselines.}
\label{sec:baselines}
We compare GFSVI to two weight-space inference methods: mean-field VI (MFVI) \citep{blundell2015weight} and linearized Laplace \citep{immer2021linlaplace}; and to three function-space inference methods: FVI \citep{sun2018functional}, TFVSI \citep{rudner2022fsvi} and VIP \citep{ma2019variational} (TFSVI performs inference in function space but with the pushforward of a weight-space prior; VIP uses a BNN prior).
All BNNs have the same architecture and fully-factorized Gaussian approximate posterior.
We also include results for a sparse GP with a posterior mean parameterized by a neural network (GWI) \citep{wild2022gvi}, and for a Gaussian Process (GP) \citep{williams2006gaussian} (when the size of the dataset allows it), and for a sparse GP \citep{hensman2013gaussian} for regression tasks.
We consider the GP, sparse~GP and GWI as gold standards as they represent the exact (or near exact) posterior for models with GP priors. 
\begin{figure*}[t]
    \centering
    \resizebox{\linewidth}{!}{
    \includegraphics[width=\linewidth]{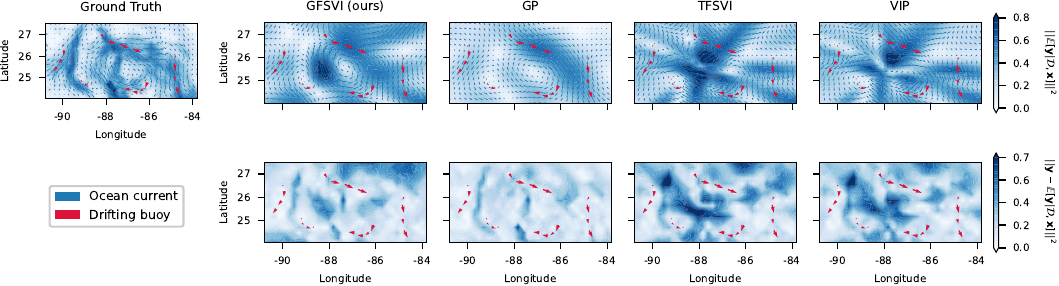}
    }
    \caption{Results for the ocean current modeling experiment. We report the norm of the mean velocity vectors and the squared errors. Unlike TFSVI, we find that GFSVI accurately captures ocean current dynamics.}
    \label{fig:ocean_current}
\end{figure*}
\paragraph{Qualitative results on synthetic data.}
\label{sec:uncert-viz}
We consider a $1$-dimensional regression task where the values~$y_i$ are sampled around $\sin(2\pi x_i)$ (circles in Figures~\ref{fig:fsvi_prior_varying_smoothness}-\ref{fig:fsvi_vs_fvi_robert} and~\ref{fig:fsvi_prior_elicitation}) and the two moons 2-dimensional binary classification task \citep{scikitlearn} (see Figures~\ref{fig:classification_gfsvi_RBF_vs_baselines} and~\ref{fig:classification_gfsvi_Matern12_vs_baselines}).
For regression, the green lines show functions sampled from the (approximate) posteriors, and the red lines are the inferred mean functions.
For classification, the first and second row show the inferred mean probability of class 1 (blue dots) and its 2-standard deviations with respect to posterior samples.
More details in Appendix~\ref{app:sec_synthetic_data}.
We find that GFSVI captures the beliefs of the RBF and Matérn-1/2 GP priors better than BNN-baselines in the regression setting (see~\cref{fig:fsvi_RBF_vs_baselines,fig:fsvi_matern_vs_baselines}) as well as in classification (see~\cref{fig:classification_gfsvi_RBF_vs_baselines,fig:classification_gfsvi_Matern12_vs_baselines}), and shows greater uncertainty outside of the support of the data.
\cref{fig:fsvi_prior_varying_smoothness,fig:fsvi_rbf_varying_lengscale} show that GFSVI notably adapts to varying prior assumptions (varying smoothness and length scale, respectively).
In addition, \cref{fig:fsvi_varying_label_noise,fig:fsvi_vs_fvi_robert} in the Appendix show that GFSVI provides strong regularization when the data generative process is noisy, and that it can be trained with fewer measurement points~$M$ than FVI without significant degradation.
\paragraph{Inductive biases.}
\cref{fig:fsvi_prior_elicitation} in the Appendix compares GFSVI to the exact GP-posterior across two different priors and three model architectures (details in Appendix~\ref{app:sec_synthetic_data}).
We find that, with ReLU activations, small models are prone to underfitting for smooth priors (RBF), and to collapsing uncertainty for rough priors (Matérn-1/2).
By contrast, with smooth activations (Tanh), smaller models suffice, and they are compatible with most standard GP priors (the results shown in \cref{fig:fsvi_prior_elicitation} extend to RBF, Matérn, and Rational Quadratic in our experiments).
We also analyzed how the number~$M$ of measurement points affects performance.
\cref{fig:fsvi_varying_n_context_points,fig:kernel_gram_eigendecay} in the appendix show that capturing the beliefs of rough GP priors and estimating $D_\text{KL}^\gamma$ with these priors requires larger~$M$.
\subsection{Quantitative results on real-world data}
\label{sec:results-quantitative}
\begin{table*}[h!]
\scshape
\caption{Test expected log-likelihood (higher is better) of evaluated methods on regression datasets. GFSVI performs competitively compared to all BNN baselines and obtains the best mean rank.
}
\label{tab:expected_ll}
\resizebox{\linewidth}{!}{
\renewcommand{\arraystretch}{1.0}
\begin{tabular}{@{}lccccccccc@{}}
\toprule
\multicolumn{1}{@{}l}{Dataset} & \multicolumn{2}{c}{Function-space priors} & \multicolumn{4}{c}{Weight-space priors} & \multicolumn{3}{c}{Gaussian Processes (Gold Standards)} \\ 
\cmidrule(rl){2-3} \cmidrule(rl){4-7} \cmidrule(rl){8-10}
\multicolumn{1}{@{}l}{} & GFSVI (ours) & \multicolumn{1}{c}{FVI} & TFSVI & MFVI & VIP & Laplace & GWI & Sparse GP & GP  \\ 
\midrule
Boston & \textbf{-0.733 $\pm$ 0.144} & \textbf{-0.571 $\pm$ 0.113} & -1.416 $\pm$ 0.046 & -1.308 $\pm$ 0.052 & \textbf{-0.722 $\pm$ 0.196} & \textbf{-0.812 $\pm$ 0.205} & -0.940 $\pm$ 0.145 & -0.884 $\pm$ 0.182 & -1.594 $\pm$ 0.556 \\
Concrete & \textbf{-0.457 $\pm$ 0.041} & \textbf{-0.390 $\pm$ 0.017} & -0.983 $\pm$ 0.012 & -1.353 $\pm$ 0.018 & \textbf{-0.427 $\pm$ 0.050} & -0.715 $\pm$ 0.025 & -0.744 $\pm$ 0.079 & -0.966 $\pm$ 0.025 & -2.099 $\pm$ 0.421 \\
Energy & \textbf{\hphantom{-}1.319 $\pm$ 0.052} & \textbf{\hphantom{-}1.377 $\pm$ 0.042} & \hphantom{-}0.797 $\pm$ 0.098 & -0.926 $\pm$ 0.197 & \textbf{\hphantom{-}1.046 $\pm$ 0.378} & \textbf{\hphantom{-}1.304 $\pm$ 0.043} & \hphantom{-}0.461 $\pm$ 0.093 & -0.206 $\pm$ 0.027 & -0.205 $\pm$ 0.022 \\
Kin8nm & -0.136 $\pm$ 0.013 & -0.141 $\pm$ 0.023 & -0.182 $\pm$ 0.011 & -0.641 $\pm$ 0.225 & \textbf{-0.102 $\pm$ 0.013} & -0.285 $\pm$ 0.014 & -0.708 $\pm$ 0.054 & -0.443 $\pm$ 0.014 & \textit{(infeasible)} \\
Naval & \textbf{\hphantom{-}3.637 $\pm$ 0.132} & \hphantom{-}2.165 $\pm$ 0.194 & \hphantom{-}2.758 $\pm$ 0.044 & \hphantom{-}1.034 $\pm$ 0.160 & \hphantom{-}1.502 $\pm$ 0.061 & \hphantom{-}3.404 $\pm$ 0.084 & -0.301 $\pm$ 0.254 & \hphantom{-}4.951 $\pm$ 0.014 & \textit{(infeasible)} \\
Power & \textbf{\hphantom{-}0.044 $\pm$ 0.011} & \textbf{\hphantom{-}0.031 $\pm$ 0.021} & \hphantom{-}0.007 $\pm$ 0.013 & -0.003 $\pm$ 0.015 & \textbf{\hphantom{-}0.036 $\pm$ 0.018} & -0.002 $\pm$ 0.019 & \hphantom{-}0.043 $\pm$ 0.009 & -0.100 $\pm$ 0.010 & \textit{(infeasible)} \\
Protein & -1.036 $\pm$ 0.005 & -1.045 $\pm$ 0.005 & -1.010 $\pm$ 0.004 & -1.112 $\pm$ 0.007 & -0.994 $\pm$ 0.007 & -1.037 $\pm$ 0.006 & -1.050 $\pm$ 0.009 & -1.035 $\pm$ 0.002 & \textit{(infeasible)} \\
Wine & -1.289 $\pm$ 0.040 & \textbf{-1.215 $\pm$ 0.007} & -2.138 $\pm$ 0.221 & -1.248 $\pm$ 0.018 & -1.262 $\pm$ 0.025 & -1.249 $\pm$ 0.025 & -1.232 $\pm$ 0.038 & -1.240 $\pm$ 0.037 & -1.219 $\pm$ 0.035 \\
Yacht & \textbf{\hphantom{-}1.058 $\pm$ 0.080} & \textbf{\hphantom{-}0.545 $\pm$ 0.735} & -1.187 $\pm$ 0.064 & -1.638 $\pm$ 0.030 & -0.062 $\pm$ 1.378 & \hphantom{-}0.680 $\pm$ 0.171 & \hphantom{-}0.441 $\pm$ 0.138 & -0.976 $\pm$ 0.092 & -0.914 $\pm$ 0.045 \\
Wave & \hphantom{-}5.521 $\pm$ 0.036 & \hphantom{-}6.612 $\pm$ 0.008 & \hphantom{-}5.148 $\pm$ 0.117 & \textbf{\hphantom{-}6.883 $\pm$ 0.008} & \hphantom{-}4.043 $\pm$ 0.093 & \hphantom{-}4.658 $\pm$ 0.027 & \hphantom{-}1.566 $\pm$ 0.123 & \hphantom{-}4.909 $\pm$ 0.001 & \textit{(infeasible)} \\
Denmark & \textbf{-0.487 $\pm$ 0.013} & -0.801 $\pm$ 0.005 & -0.513 $\pm$ 0.013 & -0.675 $\pm$ 0.007 & -0.583 $\pm$ 0.021 & -0.600 $\pm$ 0.008 & -0.841 $\pm$ 0.026 & -0.768 $\pm$ 0.001 & \textit{(infeasible)} \\
\midrule
Mean rank & 1.545 & 2.000 & 2.727 & 3.455 & 2.091 & 2.455 & - & - & - \\
\bottomrule
\end{tabular}
}
\end{table*}
\begin{table*}[h!]
\scshape
\caption{
Test expected log-likelihood, accuracy, expected calibration error and OOD detection accuracy on MNIST and Fashion MNIST.
}
\label{tab:classification}
\resizebox{\linewidth}{!}{
\renewcommand{\arraystretch}{1.}
\begin{tabular}{@{}llccccccccccc@{}}
\toprule
 & Metric & \multicolumn{4}{c}{Function-space priors} & \multicolumn{5}{c}{Weight-space priors} & GP-based \\
\cmidrule(rl){3-6} \cmidrule(rl){7-11} \cmidrule(rl){12-12} 
&  & GFSVI (rnd) & GFSVI (kmnist) & FVI (rnd) & FVI (kmnist) & TFSVI (rnd) & TFSVI (kmnist) & MFVI &VIP & Laplace & GWI \\
\midrule
\multirow{4}{*}{\parbox[t]{0pt}{\multirow{2}{*}{\rotatebox[origin=c]{90}{MNIST\hspace{-1.3em}}}}}
& Log-like.\ ($\uparrow$) & \textbf{-0.033 $\pm$ 0.000} & -0.041 $\pm$ 0.000 & -0.145 $\pm$ 0.005 & -0.238 $\pm$ 0.006 & -0.047 $\pm$ 0.003 & -0.041 $\pm$ 0.001 & -0.078 $\pm$ 0.001 & \textbf{-0.033 $\pm$ 0.001} & -0.108 $\pm$ 0.002 & -0.090 $\pm$ 0.003 \\
& Acc.\ ($\uparrow$) & \textbf{\hphantom{-}0.992 $\pm$ 0.000} & \hphantom{-}0.991 $\pm$ 0.000 & \hphantom{-}0.976 $\pm$ 0.001 & \hphantom{-}0.943 $\pm$ 0.001 & \hphantom{-}0.989 $\pm$ 0.000 & \hphantom{-}0.989 $\pm$ 0.000 & \hphantom{-}0.990 $\pm$ 0.000 & \hphantom{-}0.989 $\pm$ 0.000 & \hphantom{-}0.984 $\pm$ 0.000 & \hphantom{-}0.971 $\pm$ 0.001 \\
& ECE ($\downarrow$) & \textbf{\hphantom{-}0.002 $\pm$ 0.000} & \hphantom{-}0.006 $\pm$ 0.000 & \hphantom{-}0.064 $\pm$ 0.001 & \hphantom{-}0.073 $\pm$ 0.003 & \hphantom{-}0.007 $\pm$ 0.000 & \hphantom{-}0.006 $\pm$ 0.000 & \hphantom{-}0.021 $\pm$ 0.000 & \textbf{\hphantom{-}0.002 $\pm$ 0.001} & \hphantom{-}0.048 $\pm$ 0.001 & \hphantom{-}0.003 $\pm$ 0.000 \\
& OOD acc.\ ($\uparrow$) & \hphantom{-}0.921 $\pm$ 0.008 & \textbf{\hphantom{-}0.980 $\pm$ 0.004} & \hphantom{-}0.894 $\pm$ 0.010 & \hphantom{-}0.891 $\pm$ 0.006 & \hphantom{-}0.887 $\pm$ 0.011 & \hphantom{-}0.893 $\pm$ 0.005 & \hphantom{-}0.928 $\pm$ 0.002 & \hphantom{-}0.871 $\pm$ 0.012 & \hphantom{-}0.903 $\pm$ 0.007 & \hphantom{-}0.829 $\pm$ 0.007 \\
\midrule
\multirow{4}{*}{\parbox[t]{0pt}{\multirow{2}{*}{\rotatebox[origin=c]{90}{FMNIST\hspace{-1.5em}}}}}
& Log-like.\ ($\uparrow$) & -0.260 $\pm$ 0.003 & -0.294 $\pm$ 0.006 & -0.300 $\pm$ 0.002 & -0.311 $\pm$ 0.005 & -0.261 $\pm$ 0.001 & -0.261 $\pm$ 0.002 & -0.290 $\pm$ 0.002 & \textbf{-0.252 $\pm$ 0.001} & -0.426 $\pm$ 0.009 & -0.260 $\pm$ 0.001 \\
& Acc.\ ($\uparrow$) & \hphantom{-}0.910 $\pm$ 0.001 & \hphantom{-}0.909 $\pm$ 0.001 & \hphantom{-}0.910 $\pm$ 0.002 & \hphantom{-}0.906 $\pm$ 0.002 & \hphantom{-}0.909 $\pm$ 0.001 & \hphantom{-}0.907 $\pm$ 0.001 & \textbf{\hphantom{-}0.913 $\pm$ 0.001} & \hphantom{-}0.911 $\pm$ 0.001 & \hphantom{-}0.886 $\pm$ 0.001 & \hphantom{-}0.906 $\pm$ 0.000 \\
& ECE ($\downarrow$) & \hphantom{-}0.020 $\pm$ 0.003 & \hphantom{-}0.042 $\pm$ 0.002 & \hphantom{-}0.027 $\pm$ 0.005 & \hphantom{-}0.024 $\pm$ 0.002 & \hphantom{-}0.022 $\pm$ 0.002 & \hphantom{-}0.021 $\pm$ 0.002 & \textbf{\hphantom{-}0.010 $\pm$ 0.001} & \hphantom{-}0.024 $\pm$ 0.001 & \hphantom{-}0.060 $\pm$ 0.004 & \hphantom{-}0.016 $\pm$ 0.001 \\
& OOD acc.\ ($\uparrow$) & \hphantom{-}0.853 $\pm$ 0.005 & \textbf{\hphantom{-}0.997 $\pm$ 0.001} & \hphantom{-}0.925 $\pm$ 0.005 & \hphantom{-}0.975 $\pm$ 0.002 & \hphantom{-}0.802 $\pm$ 0.006 & \hphantom{-}0.779 $\pm$ 0.010 & \hphantom{-}0.805 $\pm$ 0.010 & \hphantom{-}0.790 $\pm$ 0.010 & \hphantom{-}0.826 $\pm$ 0.006 & \hphantom{-}0.792 $\pm$ 0.005 \\
\bottomrule
\end{tabular}
}
\end{table*}
We evaluate GFSVI on regression, classification, and out-of-distribution detection.
In all tables, we bold the highest score and any score whose error bar (standard error) overlaps with the highest score's error bar.
\begin{table}[h!]
\scshape
\centering
\caption{Results for the ocean current modeling task. 
}
\label{tab:ocean_current}
\resizebox{1\linewidth}{!}{
\renewcommand{\arraystretch}{1.}
\begin{tabular}{@{}lcccc@{}}
\toprule
\multicolumn{1}{@{}l}{Metric} & GFSVI (ours) & TFSVI & VIP & GP  \\ 
\midrule
Log-like. & -6.627 $\pm$ 0.753 & -22.651 $\pm$ 2.947 & -11.631 $\pm$ 3.171 & -0.507 $\pm$ 0.000 \\
MSE & \hphantom{-}0.021 $\pm$ 0.002 & \hphantom{-}0.034 $\pm$ 0.003 & \hphantom{-}0.026 $\pm$ 0.001 & \hphantom{-}0.013 $\pm$ 0.000\\
\bottomrule
\end{tabular}
}
\end{table}

\paragraph{Ocean current modeling.}
We measure how well GFSVI can incorporate knowledge specified via a GP prior on real-world data by considering the problem of modeling ocean currents in the Gulf of Mexico.
We follow the setup by \citet{shalashilin2024GPocean} and use the GulfDrifters dataset \citep{lilly2021GulfDrifters} to estimate ocean currents from $20$ $2$-dimensional velocity vectors collected from drifter buoys.
We embed physical properties of fluid motions into the GP prior and to the neural networks by applying the Helmholtz decomposition \citep{berlinghieri2023gaussian,cinquin2024fsplaplace}.
We compare our GFSVI to a GP, to TFSVI and to VIP.
More details can be found in \cref{app:sec_ocean_current_details}.
We find that incorporating knowledge via an informative GP prior in GFSVI improves performance over weight-space priors in TFSVI and VIP (see \cref{tab:ocean_current} and \cref{fig:ocean_current}). 
However, the GP outperforms both BNNs, which suggests that the physically motivated kernel describes the fluid dynamics well enough that the additional inductive bias introduced by a neural network hurts performance rather than helping it.
In the following, we consider experiments with larger datasets (making exact GP inference computationally infeasible in many cases), and where structural prior knowledge in function space exists but is not derived from laws of nature.
\begin{table*}[h!]
\scshape
\caption{Out-of-distribution accuracy (higher is better) of evaluated methods on regression datasets. GFSVI (ours) performs competitively on OOD detection and obtains the highest mean rank.}
\label{tab:ood_detect}
\resizebox{\linewidth}{!}{
\renewcommand{\arraystretch}{1.}
\begin{tabular}{@{}lcccccccccc@{}}
\toprule
\multicolumn{1}{@{}l}{Dataset} & \multicolumn{2}{c}{Function-space priors} & \multicolumn{4}{c}{Weight-space priors} & \multicolumn{3}{c}{Gaussian Processes (Gold Standards)} \\ 
\cmidrule(rl){2-3} \cmidrule(rl){4-7} \cmidrule(rl){8-10}
\multicolumn{1}{@{}l}{} & GFSVI (ours) & \multicolumn{1}{c}{FVI} & TFSVI & MFVI & VIP & Laplace & GWI & Sparse GP & GP  \\ 
\midrule
Boston & \textbf{0.893 $\pm$ 0.011} & 0.594 $\pm$ 0.024 & 0.705 $\pm$ 0.107 & 0.563 $\pm$ 0.013 & 0.628 $\pm$ 0.010 & 0.557 $\pm$ 0.009 & 0.817 $\pm$ 0.017 & 0.947 $\pm$ 0.011 & 0.952 $\pm$ 0.003 & \\
Concrete & \textbf{0.656 $\pm$ 0.016} & 0.583 $\pm$ 0.022 & 0.511 $\pm$ 0.003 & 0.605 $\pm$ 0.012 & 0.601 $\pm$ 0.024 & 0.578 $\pm$ 0.015 & 0.730 $\pm$ 0.020 & 0.776 $\pm$ 0.006 & 0.933 $\pm$ 0.004 & \\
Energy & \textbf{0.997 $\pm$ 0.002} & 0.696 $\pm$ 0.017 & \textbf{0.997 $\pm$ 0.001} & 0.678 $\pm$ 0.014 & 0.682 $\pm$ 0.037 & 0.782 $\pm$ 0.020 & 0.998 $\pm$ 0.001 & 0.998 $\pm$ 0.001 & 0.998 $\pm$ 0.001 & \\
Kin8nm & 0.588 $\pm$ 0.007 & \textbf{0.604 $\pm$ 0.023} & 0.576 $\pm$ 0.008 & 0.570 $\pm$ 0.009 & 0.563 $\pm$ 0.015 & \textbf{0.606 $\pm$ 0.009} & 0.602 $\pm$ 0.011 & 0.608 $\pm$ 0.014 & \textit{(infeasible)} & \\
Naval & \textbf{1.000 $\pm$ 0.000} & \textbf{1.000 $\pm$ 0.000} & \textbf{1.000 $\pm$ 0.000} & 0.919 $\pm$ 0.017 & 0.621 $\pm$ 0.059 & \textbf{1.000 $\pm$ 0.000} & 1.000 $\pm$ 0.000 & 1.000 $\pm$ 0.000 & \textit{(infeasible)} & \\
Power & \textbf{0.698 $\pm$ 0.006} & 0.663 $\pm$ 0.021 & 0.676 $\pm$ 0.008 & 0.636 $\pm$ 0.019 & 0.514 $\pm$ 0.004 & 0.654 $\pm$ 0.013 & 0.754 $\pm$ 0.004 & 0.717 $\pm$ 0.004 & \textit{(infeasible)} & \\
Protein & \textbf{0.860 $\pm$ 0.011} & 0.810 $\pm$ 0.022 & \textbf{0.841 $\pm$ 0.018} & 0.693 $\pm$ 0.020 & \textbf{0.549 $\pm$ 0.020} & 0.629 $\pm$ 0.013 & 0.942 $\pm$ 0.002 & 0.967 $\pm$ 0.001 & \textit{(infeasible)} & \\
Wine & 0.665 $\pm$ 0.013 & 0.517 $\pm$ 0.004 & 0.549 $\pm$ 0.015 & 0.542 $\pm$ 0.009 & \textbf{0.706 $\pm$ 0.028} & 0.531 $\pm$ 0.007 & 0.810 $\pm$ 0.008 & 0.781 $\pm$ 0.014 & 0.787 $\pm$ 0.007 & \\
Yacht & 0.616 $\pm$ 0.030 & 0.604 $\pm$ 0.025 & \textbf{0.659 $\pm$ 0.043} & \textbf{0.642 $\pm$ 0.035} & \textbf{0.688 $\pm$ 0.040} & 0.612 $\pm$ 0.024 & 0.563 $\pm$ 0.014 & 0.762 $\pm$ 0.018 & 0.787 $\pm$ 0.011 & \\
Wave & \textbf{0.975 $\pm$ 0.005} & 0.642 $\pm$ 0.004 & 0.835 $\pm$ 0.034 & 0.658 $\pm$ 0.026 & 0.500 $\pm$ 0.000 & 0.529 $\pm$ 0.005 & 0.903 $\pm$ 0.001 & 0.513 $\pm$ 0.001 & \textit{(infeasible)} & \\
Denmark & 0.521 $\pm$ 0.006 & \textbf{0.612 $\pm$ 0.008} & 0.519 $\pm$ 0.006 & 0.513 $\pm$ 0.003 & 0.500 $\pm$ 0.000 & 0.529 $\pm$ 0.008 & 0.688 $\pm$ 0.003 & 0.626 $\pm$ 0.002 & \textit{(infeasible)} & \\
\midrule
Mean rank & 1.455 & 2.364 & 1.909 & 2.909 & 3.364 & 2.909 & - & - & - \\
\bottomrule
\end{tabular}
}
\end{table*}
\paragraph{Regression.}
\label{sec:regression}
We assess the predictive performance of GFSVI on data sets from the UCI repository \citep{Dua2019UCI}.
\Cref{tab:expected_ll}, and \Cref{tab:mse} in the appendix, show expected log-likelihood and mean squared error, respectively.
We perform 5-fold cross validation and report means and standard errors across the test folds.
We also rank the methods for each dataset and report the mean rank of each method across all datasets.
See \cref{app:sec_regression_details} for more details.
We find that GFSVI performs competitively compared to baselines and obtains the best mean rank for both metrics, matching the top performing methods on nearly all datasets. 
In particular, we find that using GP priors in the linearized BNN with GFSVI yields improvements over the weight-space priors used in TFSVI, and that GFSVI performs slightly better than FVI despite being simpler.
Further, we find that GFSVI approximates the exact GP-posterior more accurately that FVI (see \cref{tab:var_measure_eval} and \cref{app:var_measure_eval}), and that it converges in slightly more steps than TFSVI (\cref{fig:boston_exp_ll_convergence}).
\paragraph{Classification.}
%
We further evaluate classification performance of our method on the MNIST \citep{lecun2010mnist} and FashionMNIST \citep{xiao2017FMNIST} image data sets.
We fit the models on a random subset of 90\% of the training set, use the remaining 10\% as validation data, and evaluate on the provided test split. 
We repeat with 5 different random seeds and report the mean and standard error of the expected log-likelihood, accuracy, and expected calibration error (ECE) in \cref{tab:classification}. 
For GFSVI, FVI, and TFSVI, we tested measurement points from both a uniform random (\textsc{rnd}) distribution~$\rho(\vx)$ and from \textsc{kmnist}. Details in \cref{app:sec_classification_details}.
We find that GFSVI performs competitively on MNIST, exceeding the expected log-likelihood and accuracy of top-scoring baselines and similarly to best baselines on FashionMNIST. 
GFSVI also yields well-calibrated models with low ECE.
\paragraph{Out-of-distribution detection.}
\label{sec:ood_detection}
%
We next evaluate our method by testing if its epistemic uncertainty is predictive of out-of-distribution (OOD) data.
We consider two settings: (i) with tabular data and a Gaussian likelihood \citep{malinin2021uncertGBM}, and (ii) with image data and a categorical likelihood \citep{osawa2019practical}.
We report the accuracy of classifying OOD vs.\ in-distribution (ID) data using a (learned) threshold on the predictive uncertainty.
More details in \cref{app:sec_ood_details}.
In setting~(i), GFSVI performs competitively and obtains the highest mean rank (\cref{tab:ood_detect}).
Likewise in setting~(ii), GFSVI strongly outperforms all baselines when using the \textsc{kmnist} measurement point distribution~$\rho(\vx)$ (\cref{fig:ood_detection_plot}, \cref{tab:classification,tab:influence_rho_ood_detection}).
We find that with high-dimensional image data, the choice of measurement point distribution highly influences OOD detection accuracy (see Appendix \ref{app:influence_rho_image_data} for a discussion).
In both settings, using GP priors with GFSVI rather than weight-space priors with TFSVI is beneficial, and GFSVI also improves over FVI.
GFSVI's uncertainty is also well-calibrated under distribution shift of the input features (see \cref{app:rotated_image_data}).
\section{Related work}
\label{sec:related_work}
In this section, we review related work on function-space VI with neural networks, and on approximating functions-space measures with weight-space priors.
\paragraph{Function-space inference with neural networks.}
Prior work on function-space VI in BNNs has addressed issues (i) intractable variational posterior in function space and~(ii) intractable KL divergence discussed in Section~\ref{sec:fsvi_in_bnns}.
\citet{sun2018functional} address~(i) by using implicit score function estimators, and~(ii) by replacing the supremum with an expectation.
\citet{rudner2022fsvi} address~(i) by using a linearized BNN \citep{khan2020approximate, immer2021linlaplace, maddox2021FastAdapt}, and~(ii) by replacing the supremum with a maximum over a finite set.
Other work abandons approximating the neural network's posterior and instead uses a BNN to specify a prior \citep{ma2019variational}, or deterministic neural networks as features for Bayesian linear regression \citep{ma2021funcVIspg} or the mean of a generalized sparse GP \citep{wild2022gvi}.
Unlike our more expressive GP posterior covariance, \citet{wild2022gvi} uses a simple stationary sparse GP posterior covariance (\cref{tab:classification}) which has higher sampling cost and can lead to model misspecification (\cref{fig:gfsvi_vs_gwi_periodic}).
Our work combines linearized BNNs with generalized VI, but we use the regularized KL divergence \citep{quang2019regularizedKL}, which naturally generalizes the KL divergence and allows for informative GP priors.
\paragraph{Approximating function-space measures with weight-space priors.}
\citet{FlamShepherd2017MappingGP,tran2022NeedGoodfuncpriorforBNNs} minimize a divergence between the BNN's prior predictive and a GP before performing inference on weights, while \citet{wu2023indirect} directly incorporate the bridging divergence inside the inference objective. 
Alternatively, \citet{pearce2020GPpriorsBNN} derive BNN architectures mirroring GPs, and \citet{matsubara2022ridgelet} use the Ridgelet transform to design weight-spaces priors approximating a GP in function space.
Similarly, \citet{rudner2023FuncReg} and \citet{sam2024bayesianneuralnetworksdomain} use empirical weight-space priors to regularize in function space and encode domain knowledge specified via a loss function, respectively. \citet{yang2020outputconstrainedBNN} instead imposes functional constraints directly via the prior.
\section{Discussion}
\label{sec:discussion}
We proposed a simple inference method with a well-defined variational objective function for BNNs with GP priors in function-space.
As standard VI with functions-space priors suffers from an infinite KL divergence problem, we propose to follow the generalized VI framework.
Specifically, we substitute the conventional KL divergence in the ELBO by the regularized KL divergence, which is always finite, and which can be estimated consistently within the linearized BNN approximation.
We demonstrated that our method allows to incorporate interpretable structural properties via a GP prior, accurately approximates the true GP posterior on synthetic and small real-world data sets, and provides competitive uncertainty estimates for regression, classification and out-of-distribution detection compared to BNNs with both function-space and weight-space priors.
%
Future work should investigate the use of more expressive variational distributions, such as Gaussian with low-rank plus diagonal covariance proposed by \citet{tomczak2020lowrankGausVI}.


\begin{acknowledgements}
This work was funded by the Deutsche Forschungsgemeinschaft (DFG, German Research Foundation) under Germany’s Excellence Strategy~--~EXC number 2064/1~--~Project number 390727645, and by the German Research Foundation (DFG) under project 448588364 of the Emmy Noether Programme.
Additional support was provided by the German Federal Ministry of Education and Research (BMBF): Tübingen AI Center, FKZ:~01IS18039A.
The authors extend their gratitude to the International Max Planck Research School for Intelligent Systems (IMPRS-IS) for supporting Tristan Cinquin.
Finally, Tristan Cinquin thanks Marvin Pförtner for very useful discussions on the theory and experiments, and Vincent Fortuin for feedback. 
\end{acknowledgements}

\bibliography{uai2025}

\newpage

\onecolumn

\title{Supplementary Material}
\maketitle
\appendix

\section{Divergences between Gaussian measures}
\subsection{The KL divergence is infinite}
\label{sec:infinite_kl}
In this section, we show that the Kullbach-Liebler (KL) divergence between the Gaussian measures $\measureQ_\phi^F \sim \gaussian{m_Q}{C_Q}$ and $\measureP^F\sim \gaussian{m_P}{C_P}$, respectively induced by the linearized BNN in Eq~\ref{eq:linearized-q} and by a non-degenerate Gaussian process satisfying conditions given in \cref{sec:reg_kl}, is infinite.
While this has already been shown by \citet{burt2020understanding}, the proof is easier for Gaussian measures. 
We first need the Feldman-Hàjek theorem which tells us when the KL divergence between two Gaussian measures is well-defined. 
\begin{theorem}[Feldman-Hàjek, \citet{quang2022gpkl} Theorem 2, \citet{simpson2022} Theorem 7]\label{th:feldman_hajek}
Consider two Gaussian measures $\nu_1 = \gaussian{m_1}{C_1}$ and $\nu_2 = \gaussian{m_2}{C_2}$ on $\Ltwo{\mathcal{X}}{\rho}$. Then $\nu_1$ and $\nu_2$ are called equivalent if and only if the following holds:
\begin{enumerate}[noitemsep,topsep=0pt,parsep=0pt,partopsep=0pt,leftmargin=*]
    \item $m_1 - m_2 \in \text{Im}(C_2^{1/2})$
    \item The operator $T$ such that $C_1 = C_2^{1/2}(I-T)C_2^{1/2}$ is Hilbert-Schmidt, that is $T$ has a countable set of eigenvalues $\lambda_i$ that satisfy $\lambda_i < 1$ and $\sum_{i=1}^{\infty} \lambda_i^2 < \infty$.
\end{enumerate}
otherwise $\nu_1$ and $\nu_2$ are singular.
If $\nu_1$ and~$\nu_2$ are equivalent, then the Radon-Nikodym derivative exists and $\kl{\nu_1}{\nu_2}$ admits an explicit formula. Otherwise, $\kl{\nu_1}{\nu_2} = \infty$.
\end{theorem}
Let us now show that the KL divergence between $\measureQ_\phi^F$ and $\measureP^F$ is indeed infinite.
\begin{prop}\label{prop:infinit-kl}
The Gaussian measures $\measureQ_\phi^F$ and $\measureP^F$ are mutually singular and $D_\text{KL}(\measureQ_\phi^F || \measureP^F) = \infty$.
\end{prop}
\begin{proof}
The proof follows from the Feldman-Hàjek theorem (\cref{th:feldman_hajek}). 
In our case, $C_Q$ has at most $p$ non-zero eigenvalues as the covariance function of the GP induced by the BNN is degenerate, while $C_P$ has a set of (countably) infinite non-zeros eigenvalues (prior is non-degenerate as per assumption).
Hence, for the equality in condition (2) to hold, $T$ must have eigenvalue $1$ which violates the requirement that $T$ is Hilbert-Schmidt \ie that its eigenvalues $\set{\lambda_i}_{i=1}^{\infty}$ satisfy $\lambda_i < 1$ and $\sum_{i=1}^{\infty} \lambda_i^2 < \infty$.
Therefore, $\measureQ_\phi^F$ and $\measureP^F$ are mutually singular and $D_\text{KL}(\measureQ_\phi^F || \measureP^F) = \infty$.
\end{proof}
\subsection{The regularized KL divergence}
\label{sec:app_reg_kl}
We provide the bound describing the asymptotic convergence of the regularized KL divergence estimator in \cref{eq:reg_kl_bound}. 
The error results from the fact that taking a finite number M of context points effectively cuts off the spectra of the covariance operators and the estimator $\hat{D}_{\text{KL}}^\gamma$ converges to $D_{\text{KL}}^\gamma$ as $M\to\infty$ with high probability. \\\\
\begin{theorem}[Convergence of estimator, \citet{quang2022gpkl} Theorem 45]
Assume the following:
\begin{enumerate}[noitemsep,topsep=0pt,parsep=0pt,partopsep=0pt,leftmargin=*]
    \item Let T be a $\sigma-\text{compact}$ metric space, that is $T = \cup_{i=1}^{\infty} T_i$, where $T_1 \subset T_2 \subset \cdots$ with each $T_i$ being compact.
    \item $\rho$ is a non-degenerate Borel probability measure on T, that is $\rho\br{B} > 0$ for each open set $B \subset T$.
    \item $K_1, K_2 : T \times T \to \real$ are continuous, symmetric, positive definite kernels and there exists $\kappa_1 > 0, \kappa_2 > 0$ such that $\int_T K_i(x, x) d\rho\br{x} \leq \kappa_i^2 \;$ for $i=1,2$.
    \item $\operatorname{sup}_{x \in T} K_i(x, x) \leq \kappa_i^2 \;$ for $i=1,2$.
    \item $f_i \sim GP\br{\mu_i, K_i}$, where $\mu_i \in L^2(T, \rho)$ for $i=1,2$.
    \item  $\exists B_i > 0$ such that $\norm{\mu_i}_{\infty} \leq B_i$ for $i=1,2$.
\end{enumerate}
Let $\vx = \{\vx^{(i)}\}_{i=1}^M$, $\vx^{(1)}, \dots, \vx^{(M)} \oset[.40ex]{\textup{\tiny i.i.d}}{\sim} \rho(\vx)$. If Gaussian measures $\gaussian{m_i}{C_i}$ are induced by GPs $f_i \sim \mathcal{GP}(\mu_i, K_i)$ for $i=1,2$, then for any $0 < \delta < 1$, with probability at least $1 - \delta$,
\begin{multline}\label{eq:reg_kl_bound}
    |\kl{\gaussian{\mu_1(\mathbf{x})}{K_1(\mathbf{x},\mathbf{x})+M \gamma \eye_M}}{{\gaussian{\mu_2(\mathbf{x})}{K_2(\mathbf{x}, \mathbf{x})+M \gamma \eye_M}}} 
    - \regkl{\gaussian{m_1}{C_1}}{\gaussian{m_2}{C_2}}| \\
\leq \frac{1}{2\gamma}(B_1 + B_2)^2 [1 + \kappa_2^2 / \gamma]^2 \br{\frac{2 \log \frac{48}{\delta}}{M}+\sqrt{\frac{2 \log \frac{48}{\delta}}{M}}} \\
+ \frac{1}{2\gamma^2} [\kappa_1^4 + \kappa_2^4 + \kappa_1^2 \kappa_2^2 (2 + \kappa_2^2 / \gamma)] \br{\frac{2 \log \frac{12}{\delta}}{M}+\sqrt{\frac{2 \log \frac{12}{\delta}}{M}}} 
\end{multline}
\end{theorem}
Note that \cref{eq:reg_kl_bound} provides a very general bound on the error that does not make assumptions on the spectral decay, and it may therefore dramatically overestimate the error. Indeed, we analyze convergence empirically in \cref{fig:regkl_vs_gamma} and observe that the estimator converges quickly except for very rough priors (e.g., Matérn-1/2) with very small $\gamma$.
\section{Additional details on the GFSVI objective estimator}
\label{app:sec_gfsvi_estimator}

In this section, we present details on the estimation of the generalized function-space variational inference (GFSVI) objective.
Let $f_L({\,\cdot\,; \vw})$ be the linearized BNN (Eq~\ref{eq:linearized_nn}) with weights $\vw \in \real^p$, and $\mathcal{D} = \{(\vx_i, y_i)\}_{i=1}^N$ a data set with features~$\vx_i \in \mathcal{X} \subset \real^d$ and associated values~$y_i \in \mathcal{Y}$.
Assuming a likelihood $\prob{\mathcal{D} \g \vw} = \prod_{i=1}^N \prob{y_i \g f(\vx_i; \vw)}$ and a Gaussian variational distribution on model weights $q_\phi(\vw)=\gaussianx{\vw}{\vm}{\mS}$, the GFSVI objective function is 
\begin{equation}\label{app:objective}
    \mathcal{L}(\phi) = \sum_{i=1}^N \expect{\log \prob{y_i \g f_L(\vx_i; \vw)}}{q_\phi(\vw)} - \regkl{\measureQ_\phi^F}{\measureP^F}
\end{equation}
where $\measureQ_\phi^F$ and $\measureP^F$ are the Gaussian measures induced by the linearized BNN and a Gaussian process prior respectively.

\paragraph{Expected log-likelihood}

When considering a Gaussian likelihood, we use the closed form expression available due to the Gaussian variational measure over functions induced by the linearized BNN
\begin{equation} \label{eq:objective_expected_ll_gaussian}
    \expect{\log \gaussianx{y_i}{f_L(\vx_i;\vw)}{\sigma_y^2}}{q_\phi(\vw)} = -\frac{1}{2}\log \br{2\pi \sigma_y^2} - \frac{\br{y_i - f\br{\vx_i; \vm}}^2 + J\br{\vx_i;\vm} \mS J\br{\vx_i;\vm}\tp}{2\sigma_y^2}.
\end{equation}
When considering a Categorical likelihood with $C$ different classes, we estimate the expected log-likelihood term using Monte-Carlo integration as
\begin{equation}\label{eq:objective_expected_ll_categorical}
    \expect{\log \text{Cat}(y_i \g \sigma(f_L(\vx_i;\vw)))}{q_\phi(\vw)} = \frac{1}{K} \sum_{k=1}^K \sum_{c=1}^C \eye[y_i = c] \left[ f^c_L(\vx_i;\vw^{(k)}) - \log \left[ \sum_{c'=1}^C \exp \left( f^{c'}_L(\vx_i;\vw^{(k)})\right) \right] \right]
\end{equation}
where $\vw^{(k)} \sim q_\phi(\vw)$ for $k=1, \dots, K$, $\eye[\cdot]$ is the indicator function, $\sigma(\cdot)$ is the softmax function and $f^c_L(\,\cdot\,;\vw)$ is the logit for class $c$ obtained from $f_L$. 

\paragraph{Regularized KL divergence}
We estimate the regularized KL divergence using its consistent estimator (Eq.~\ref{eq:reg-kl-estimator})
\begin{multline}\label{eq:gfsvi-reg-kl-estimator}
    \regklhat{\measureQ_\phi^F}{\measureP^F} = \frac{1}{2} \br{f(\vx; \vm) - \mu(\vx)}\tp (K(\vx, \vx)+\gamma M \eye_M)^{-1} \br{f(\vx; \vm) - \mu(\vx)} \\
    + \frac{1}{2} \operatorname{Tr}\big[(K(\vx, \vx)+\gamma M \eye_M)^{-1}(J(\vx; \vm)SJ(\vx; \vm)\tp+\gamma M \eye_M) - \eye_M\big] \\
     - \frac{1}{2} \log \det \big[(K(\vx, \vx)+\gamma M \eye_M)^{-1}(J(\vx; \vm)SJ(\vx; \vm)\tp+\gamma M \eye_M)\big]
\end{multline}
with measurement points $\vx = \{\vx^{(i)}\}_{i=1}^M$, $\vx^{(1)}, \dots, \vx^{(M)} \oset[.40ex]{\textup{\tiny i.i.d}}{\sim} \rho(\vx)$ sampled from a probability measure on $\mathcal{X}$.

\paragraph{Computational complexity}

Evaluating the objective in Eq.~\ref{app:objective} has complexity $O(BKC + M^3)$ for Categorical likelihoods and $O(B + M^3)$ for Gaussian likelihoods, where $B$ is the batch size, $K$ the number of variational posterior samples, $C$ the number of classes, and $M$ the number of context points. The first term corresponds to the expected log-likelihood in our objective and the second term to the regularized KL divergence estimator. We note that evaluating the linearized neural network can be efficiently done in about 3x the cost of one forward pass using the Jacobian-vector product computational primitive.

\section{Additional details on the experimental setup}
\label{app:sec_exp_details}

\subsection{Experiments on synthetic data}
\label{app:sec_synthetic_data}

\paragraph{Regression}
We consider the following generative model for the toy data
\begin{equation}
    y_i = \operatorname{sin}(2\pi x_i) + \epsilon \quad \text{with} \quad \epsilon \sim \gaussian{0}{\sigma_n^2}
\end{equation}
and draw $x_i \sim \mathcal{U}([-1, -0.5] \cup [0.5, 1])$.
When not otherwise specified, we use $\sigma_n = 0.1$. 
On the plots, the data points are shown as gray circles, inferred mean functions as red lines, their 2-standard-deviations interval around the mean in light green, and functions sampled from the approximate posterior as green lines.
In general, we consider two hidden-layer BNNs with 30 neurons per layer and hyperbolic tangent activation (Tanh) functions. 
Specifically in \cref{fig:fsvi_prior_elicitation}, the small BNN has the same architecture as above while the large BNN has 100 neurons per layer. 
All the BNN baselines have the same architecture and fully-factorized Gaussian approximate posterior.
The prior scale of TFSVI \citep{rudner2022fsvi} is set to $\sigma_p = 0.2$ and $\sigma_p = 0.75$ for MFVI \citep{blundell2015weight} and Laplace \citep{immer2021linlaplace}. 
For the Gaussian process posterior baseline, we fit the prior parameters by maximizing the log-marginal likelihood \citep{williams2006gaussian}.
Apart from the cases where the parameters of the GP prior used for GFSVI (our method) and FVI \citep{sun2018functional} are explicitly stated, we consider a constant zero-mean function and find the parameters of the covariance function by maximizing the log-marginal likelihood from mini-batches \citep{chen2021gaussian}.
Except where otherwise stated, we estimate the functional KL divergences with 500 measurement points and use the regularized KL divergence with $\gamma = 10^{-10}$. 

\paragraph{Classification}
We sample 100 data points perturbed by Gaussian noise with $\sigma_n = 0.1$ from the two moons data \citep{scikitlearn}.
On the plots, the data points are shown as red (class 0) and blue (class 1) dots. 
We plot the mean and 2-standard-deviations of the probability that $\vx$ belongs to class 1 with respect to the posterior  (\ie $p(y=1 \g \vw^{(k)}, \vx)$) which we estimate from $K=100$ samples $\vw^{(k)} \sim q_\phi(\vw)$ for $k= 1, \dots, K$. 
We consider two hidden-layer BNNs with 100 neurons per layer and hyperbolic tangent activation (Tanh) functions. 
All the BNN baselines have the same architecture and fully-factorized Gaussian approximate posterior.
The prior scale of MFVI \citep{blundell2015weight} is set to $\sigma_p = 0.8$ and $\sigma_p = 1.0$ for TFSVI \citep{rudner2022fsvi} and Laplace \citep{immer2021linlaplace}. 
For the Gaussian process posterior baseline, we approximate the intractable posterior using the Laplace approximation and find the prior parameters by maximizing the log-marginal likelihood \citep{williams2006gaussian}.
The GP prior for GFSVI (our method) and FVI \citep{sun2018functional} has a constant zero-mean function and we find the parameters of the covariance function by maximizing the log-marginal likelihood from mini-batches \citep{chen2021gaussian} using the method to transform classifications labels into regression targets from \citet{milios2018DirGP}.
We estimate the functional KL divergences with 500 measurement points and use the regularized KL divergence with $\gamma = 10^{-10}$. 

\subsection{Ocean current modeling experiment}
\label{app:sec_ocean_current_details}

Following \citet{cinquin2024fsplaplace}, we apply the Helmholtz decomposition to the neural network $f$ as
\begin{equation}
    f(\cdot, \vw) = \operatorname{grad} \Phi(\cdot, \vw_1) + \operatorname{rot} \Psi(\cdot, \vw_2)
\end{equation}
where $\vw = \set{\vw_1, \vw_2}$ and, $\Phi(\cdot, \vw_1)$ and $\Psi(\cdot, \vw_2)$ are 2-layer fully-connected neural networks with $50$ hidden units per layer and hyperbolic tangent activation functions. 
GFSVI and TFSVI both use $160$ fixed context points.
The prior scale of TFSVI is set to $\sigma_p=0.5$.
We fit the neural networks on the entire dataset and average the scores with respect to five different random seeds.

\subsection{Regression experiments with tabular data}
\label{app:sec_regression_details}

\paragraph{Datasets and pre-processing}
We evaluate the predictive performance of our model on regression datasets from the UCI repository \citep{Dua2019UCI} described in \cref{tab:regression_dataset_description}.
These datasets are also considered in \citet{sun2018functional, wild2022gvi} but we include two additional larger ones (Wave and Denmark).
We perform 5-fold cross validation, leave out one fold for testing, consider 10\% of the remaining 4 folds as validation data and the rest as training data.
We report mean and standard-deviation of the average expected log-likelihood and average mean square error on the test fold. 
We also report the mean rank of the methods across all datasets by assigning rank 1 to the best scoring method as well as any method who's error bars overlap with the highest score's error bars, and recursively apply this procedure to the methods not having yet been assigned a rank. 
The expected log-likelihood is estimated by Monte Carlo integration when it is not available in closed form (MFVI, TFSVI and FVI) with 100 posterior samples.
We preprocess the dataset by encoding categorical features as one-hot vectors and standardizing the features and labels. 

\paragraph{Baseline specification}
We compare our GFSVI method to two weight-space inference methods (mean-field variational inference \citep{blundell2015weight} and linearized Laplace \citep{immer2021linlaplace}) and two function-space inference methods (FVI \citep{sun2018functional} and TFSVI \citep{rudner2022fsvi}).
While FVI uses GP priors, TFSVI performs inference in function space but with the pushforward to function space of the variational distribution and prior on the weights.
We compute the function-space (regularized) KL divergence using a set of 500 measurement points sampled from a uniform distribution for GFSVI and TFSVI, and 50 points drawn from a uniform distribution along with 450 samples from the training batch for FVI as specified in \citet{sun2018functional}.
All the BNN baselines have the same architecture and fully-factorized Gaussian approximate posterior.
We also provide results with a GP \citep{williams2006gaussian} when the size of the dataset allows it, and a sparse GP \citep{hensman2013gaussian}.
As we restrict our comparison to BNNs, we do not consider the GP and sparse GP as baselines but rather as gold-standards. 
All models have a Gaussian homoskedastic noise model with a learned scale parameter.
All the BNNs are fit using the Adam optimizer \citep{kingma2017adam} using a mini-batch size of 2000 samples. 
We also perform early stopping when the validation loss stops decreasing.

\paragraph{Model selection}
Hyper-parameter optimization is conducted using the Bayesian optimization tool provided by Wandb \citep{wandb}.
BNN parameters are selected to maximize the average validation expected log-likelihood across the 5 cross-validation folds. 
We optimize over prior parameters (kernel and prior scale), learning-rate and activation function. 
We select priors for GFSVI, FVI, sparse GP and GP among the RBF, Matérn-1/2, Matérn-3/2, Matérn-5/2, Linear and Rational Quadratic covariance functions.
The GP prior parameters used with GFSVI and FVI are selected by maximizing the log-marginal likelihood from batches as proposed by \citet{chen2021gaussian} and done in \citet{sun2018functional}.
Hyper-parameters for GPs and sparse GPs (kernel parameters and learning-rate) are selected to maximize the mean log-marginal likelihood of the validation data across the 5 cross-validation folds. 

\begin{table}[t]
\scshape
\caption{UCI regression dataset description}
\label{tab:regression_dataset_description}
\resizebox{\linewidth}{!}{
\begin{tabular}{@{}llllllllllll@{}}
\toprule
Dataset & Boston & Naval  & Power & Protein & Yacht & Concrete & Energy & Kin8nm & Wine & Wave & Denmark\\ \midrule
Number samples & 506 & 11\,934 & 9\,568 & 45\,730 & 308 & 1\,030 & 768 & 8\,192 & 1\,599 & 288\,000 & 434\,874 \\
Number features & 13 & 16 & 4 & 9 & 6 & 8 & 8 & 8 & 11 & 49 & 2 \\ \bottomrule
\end{tabular}
}
\end{table}

\subsection{Classification experiments with image data}
\label{app:sec_classification_details}

\paragraph{Datasets and pre-processing}
We further evaluate the predictive performance of our model on classification tasks with the MNIST \citep{lecun2010mnist} and Fashion MNIST \citep{xiao2017FMNIST} image data sets.
We fit the models on a random subset of 90\% of the provided training split, consider the remaining 10\% as validation data and evaluate on the provided test split. 
We repeat this procedure 5 times with different random seeds and report the mean and standard-deviation of the average expected log-likelihood, accuracy and expected calibration error (ECE) of the mean of the predictive distribution on the test set.
The expected log-likelihood is estimated by Monte Carlo integration with 100 posterior samples when it is not available in closed form (MFVI, TFSVI and FVI). 
We estimate the mean of the predictive distribution to compute the accuracy and the ECE with 100 posterior samples. 
We preprocess the dataset by standardizing the images. 

\paragraph{Baseline specification}
We compare our GFSVI method to the same baselines as for the regression experiments (see \ref{app:sec_regression_details}).
All the BNN baselines have the same architecture and fully-factorized Gaussian approximate posterior.
More specifically, we consider a CNN with three convolutional layers (with output channels 16, 32 and 64) before two fully connected layers (with output size 128 and 10).
The convolutional layers use $3\times3$ shaped kernels. 
Each pair of convolutional layers is interleaved with a max-pooling layer. 
We consider three different measurement point distributions $\rho$ to estimate the (regularized) KL divergence in GFSVI, FVI and TFSVI: \textsc{random}, \textsc{random pixel} and \textsc{kmnist}.
The \textsc{random} measurement point distribution is sampled from by drawing 50\% of the samples from the training data batch and 50\% of the samples from a uniform distribution over $[p_{min}, p_{max}]^{H \times W \times C}$, where $H$, $W$ and $C$ are respectively the height, width and number of channels of the images, and $p_{min}=v_{min}-0.5 \times \Delta$ and $p_{max}=v_{max}+0.5 \times \Delta$ where $\Delta=v_{max}-v_{min}$ is the difference between the minimal ($v_{min}$) and maximal ($v_{max}$) pixel values of the data set.
The \textsc{random pixel} measurement point distribution is taken from \citet{rudner2022fsvi} and is sampled from by randomly choosing each pixel value among the ones available from the training data batch at the same position in the $28 \times 28$ pixel grid.
Finally, the \textsc{kmnist} measurement point distribution is also taken from \citet{rudner2022fsvi} and is drawn from by randomly sampling data points from the Kuzushiji-MNIST (KMNIST) dataset \citep{clanuwat2018kmnist}. 
The KMNIST dataset is a collection of 70'000 gray-scale images of size $28 \times 28$ which we preprocess by standardizing the images.
We sample 25 measurement points when using \textsc{random}, 25 measurement points when using \textsc{random pixel} and 20 when using \textsc{kmnist}.
All the BNNs are trained using the Adam optimizer \citep{kingma2017adam} using a mini-batch size of 100. 
We also perform early stopping when the validation loss stops decreasing.

\paragraph{Model selection}
Hyper-parameter optimization is conducted just like for the regression tasks (see \ref{app:sec_regression_details}).
The Gaussian process prior parameters used with GFSVI and FVI are selected by maximizing the log-marginal likelihood from batches \citep{chen2021gaussian} using the method to transform classifications labels into regression targets from \citet{milios2018DirGP}.
We optimize the same hyper-parameters as for the regression experiments with the exception of the additional $\alpha_\epsilon$ parameter introduced by \citet{milios2018DirGP} for the function-space VI methods with GP priors (FVI and GFSVI).

\subsection{OOD detection}
\label{app:sec_ood_details}

\paragraph{Tabular data with a Gaussian likelihood}

Following the setup from \citet{malinin2021uncertGBM} we take epistemic uncertainty to be the variance of the mean prediction with respect to samples from the posterior.
We consider the test data to be in-distribution (ID) data and a subset of the song dataset \citep{Bertin-Mahieux2011} of equal length and with an equal number of features as out-of-distribution (OOD) data.
We use the same preprocessing as for regression as well as the same baselines with the same hyper-parameters (see \cref{app:sec_regression_details}).
We first fit a model, then evaluate the extend by which the epistemic uncertainty under the model is predictive of the ID and OOD data using a single threshold obtained by a depth-1 decision tree fit to minimize the classification loss.
We report the mean and standard error of the accuracy of the threshold to classify OOD from ID data based on epistemic uncertainty across the 5 folds of cross-validation.
We also provide results obtained using a GP and sparse GP as gold standard.

\paragraph{Image data with a Categorical likelihood}

Following the setup by \citet{osawa2019practical}, we take the epistemic uncertainty to be the entropy of the mean of the predictive distribution with respect to samples from the posterior. 
We evaluate models trained on MNIST using MNIST's test split as ID data and a subset of the training set of Fashion MNIST as OOD data. 
Likewise, we evaluate models trained on Fashion MNIST using Fashion MNIST's test split as ID data and a subset of the training set of MNIST as OOD data. 
We use the same preprocessing as for classification, as well as the same baselines with the same hyper-parameters (see \cref{app:sec_classification_details}).
We first fit a model, then evaluate the extend by which the epistemic uncertainty under the model is predictive of the ID and OOD data using a single threshold obtained by a depth-1 decision tree fit to minimize the classification loss.
We estimate mean of the predictive distribution by Monte-Carlo integration using 100 posterior samples.
We report the mean and standard error of the accuracy of the threshold to classify OOD from ID data based on epistemic uncertainty for the 5 models trained on different random seeds (see \cref{app:sec_classification_details}).

\subsection{Variational measure evaluation}
\label{app:sec_measure_eval_details}

We evaluate our inference method by comparing the samples drawn from the exact posterior over functions with the approximate posterior obtained with our method (GFSVI).
We follow the setup by \citet{wilson2022evalBNN} and we compute the average Wasserstein-2 metric between 1000 samples drawn from a GP posterior with a RBF kernel evaluated at the test points, and samples from the approximate posterior of GFSVI, sparse GP and FVI evaluated at the same points and with the same prior. 
We consider the Boston, Concrete, Energy, Wine and Yacht datasets for which the exact GP posterior can be computed and use the same preprocessing as for regression (see \cref{app:sec_regression_details}).
We report the mean and standard error of the average Wasserstein-2 metric across the 5 folds of cross-validation.
The Wasserstein-2 metric is computed using the Python Optimal Transport library \citep{flamary2021pot}.

\paragraph{Baseline specification}
FVI and GFSVI have the same two hidden layer neural network architecture with 100 neurons each and hyperbolic tangent activation.
These models are fit with the same learning rate and set of 500 measurement points jointly sampled from a uniform distribution over the feature-space and mini-batch of training samples.
We use $\gamma=10^{-15}$ for the regularized KL divergence. 
We further consider a sparse GP with 100 inducing points.

\subsection{Software}
We use the JAX \citep{jax2018github} and DM-Haiku \citep{haiku2020github} Python libraries to implement our Bayesian neural networks.
MFVI, linearized Laplace and TFSVI were implemented based on the information in the papers, and code for FVI was adapted to the JAX library from the implementation provided by the authors. 
We further use the GPJAX Python library for experiments involving Gaussian processes \citep{Pinder2022gpjax}. 

\subsection{Hardware}
All models were fit using a single NVIDIA RTX 2080Ti GPU with 11GB of memory.

\section{Additional experimental results}

In this section, we present additional figures for our qualitative uncertainty evaluation as well as further experimental results on regression, out-of-distribution detection and robustness under input distribution shift tasks.
We also provide plots illustrating the eigenvalue decay of different kernels, and figures showing the influence of $\gamma$ in the regularized KL divergence.

\subsection{Qualitative uncertainty evaluation}

\begin{figure*}[t]
    \centering
    \resizebox{\linewidth}{!}{
    \includegraphics[width=\linewidth]{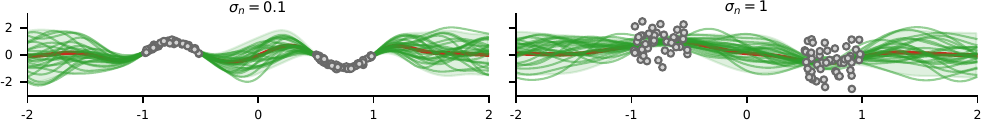}
    }
    \caption{Our method (GFSVI) effectively regularizes functions generated by the Bayesian neural network (BNN) both in settings where the generative process is very noisy ($\sigma_n=1$) or not ($\sigma_n=0.1$).}
    \label{fig:fsvi_varying_label_noise}
\end{figure*}

\paragraph{Regression}

\begin{figure*}[!h]
    \centering
    \resizebox{\linewidth}{!}{
    \includegraphics[width=\linewidth]{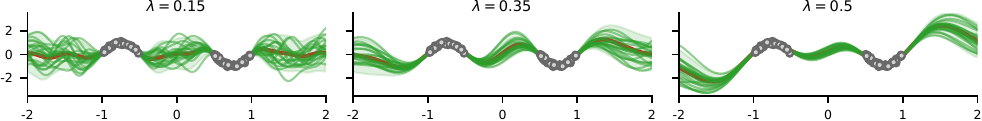}
    }
    \caption{Our method (GFSVI) allows to incorporate prior beliefs in terms of function variability using the characteristic length-scale parameter of the Gaussian process (GP) prior. GFSVI was fit using a GP prior with RBF covariance function.}
    \label{fig:fsvi_rbf_varying_lengscale}
\end{figure*}

\begin{figure*}[!h]
    \centering
    \resizebox{\linewidth}{!}{
    \includegraphics[width=\linewidth]{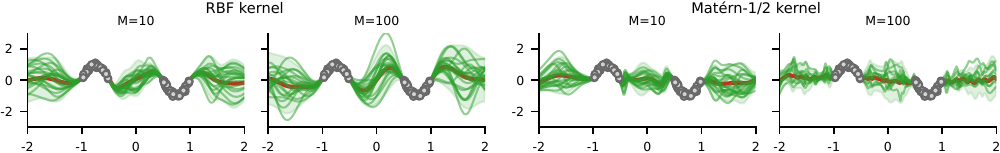}
    }
    \caption{Our method (GFSVI) captures the smooth behavior of a Gaussian process (GP) prior with RBF covariance function even if the number of measurement points is small (M=10). However, in that setting GFSVI fails to reproduce the rough effect of a GP prior with a Matérn-1/2 covariance function, and requires a larger amount of measurement points to do so (M=100).}
    \label{fig:fsvi_varying_n_context_points}
\end{figure*}

\begin{figure*}[!h]
    \centering
    \resizebox{\linewidth}{!}{
    \includegraphics[width=\linewidth]{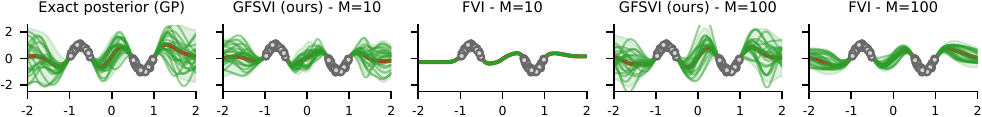}
    }
    \caption{Our method (GFSVI) already provides a reasonable approximation to the exact posterior with small numbers of measurement points (M=10) while function-space baseline FVI requires many more (M=100).}
    \label{fig:fsvi_vs_fvi_robert}
\end{figure*}

We further find that our method (GFSVI) provides strong regularization when the data generative process is noisy (see \cref{fig:fsvi_varying_label_noise}) and is more robust than FVI to situations where ones computational budget constrains the number of measurement points $M$ to be small (\cref{fig:fsvi_vs_fvi_robert}).
In contrast to FVI, GFSVI accurately approximates the exact GP posterior under rough (Matérn-1/2) GP priors effectively incorporating prior knowledge defined by the GP prior to the inference process (see \cref{fig:fsvi_matern_vs_baselines}).
Likewise, GFSVI adapts to the variability of the functions specified by the kernel (see \cref{fig:fsvi_rbf_varying_lengscale}).
We also find that GFSVI requires a larger number of measurement points to capture the behavior of a rougher prior (see \cref{fig:fsvi_varying_n_context_points}).

\begin{figure*}[t]
    \centering
    \resizebox{\linewidth}{!}{
    \includegraphics[width=\linewidth]{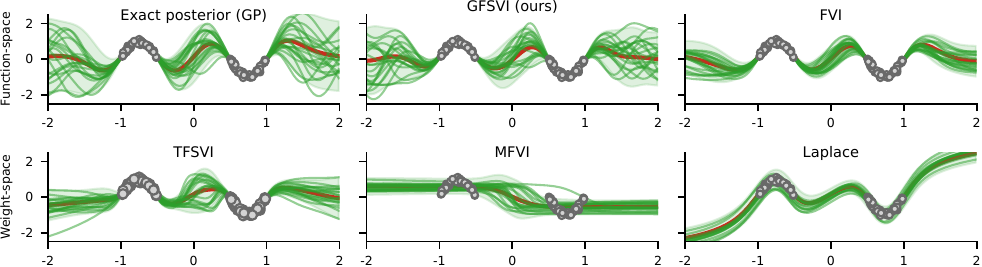}
    }
    \caption{Our method (GFSVI) with an RBF Gaussian process (GP) prior accurately approximates the exact GP posterior unlike the function-space prior baseline (FVI). Weight-space prior baselines do not provide a straight-forward mechanism to incorporate prior assumptions regarding the functions generated by BNNs and underestimate the epistemic uncertainty (MFVI, Laplace). The lower row is identical to the one in \cref{fig:fsvi_matern_vs_baselines} in the main text and is reproduced here to make comparison easier.}
    \label{fig:fsvi_RBF_vs_baselines}
\end{figure*}

\paragraph{Classification}

\begin{figure*}[!h]
    \centering
    \resizebox{\linewidth}{!}{
    \includegraphics[width=\linewidth]{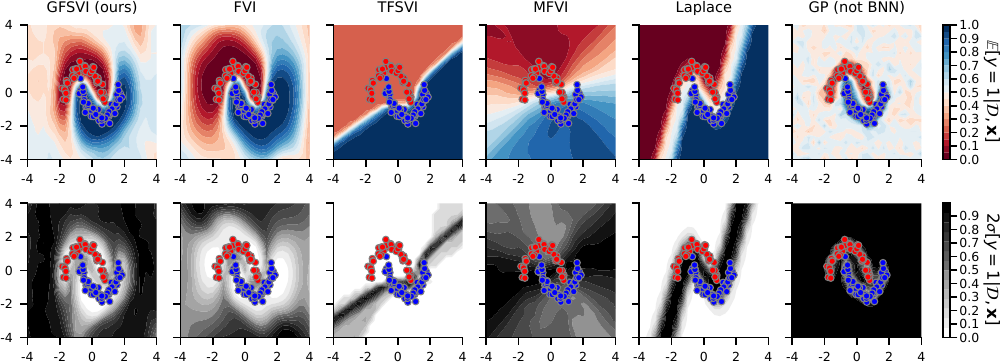}
    }
    \caption{Our method (GFSVI) with a RBF Gaussian process (GP) prior accurately captures the smooth decision boundary induced by the prior and shows high uncertainty outside of the data support. 
    Weight-space baselines do not provide a straight-forward mechanism to incorporate prior assumptions regarding the functions generated by BNNs and underestimate the epistemic uncertainty (TFSVI, Laplace) or underfit the data (MFVI).}
    \label{fig:classification_gfsvi_RBF_vs_baselines}
\end{figure*}

\begin{figure*}[!h]
    \centering
    \resizebox{\linewidth}{!}{
    \includegraphics[width=\linewidth]{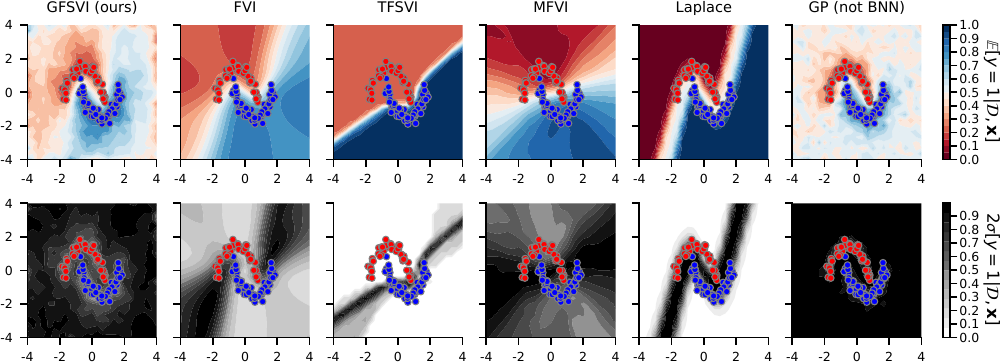}
    }
    \caption{Our method (GFSVI) with a Matérn-1/2 Gaussian process (GP) prior accurately captures the rough decision boundary unlike the function-space baseline (FVI).
    Weight-space baselines do not provide a straight-forward mechanism to incorporate prior assumptions regarding the functions generated by BNNs and underestimate the epistemic uncertainty (TFSVI, Laplace) or underfit the data (MFVI).}
    \label{fig:classification_gfsvi_Matern12_vs_baselines}
\end{figure*}

We find that GFSVI better captures the beliefs induced by the smooth RBF and rough Matérn-1/2 Gaussian process priors compared to FVI (see Figures~\ref{fig:classification_gfsvi_RBF_vs_baselines} and \ref{fig:classification_gfsvi_Matern12_vs_baselines}).
Moreover, GFVSI both accurately fits the training data and shows greater uncertainty outside of its support relative to BNNs baselines with weight-space and function-space priors.
Unlike for the toy data regression experiments where the GP posterior was the ground truth, the Laplace (approximate) GP posterior in Figures~\ref{fig:classification_gfsvi_RBF_vs_baselines} and \ref{fig:classification_gfsvi_Matern12_vs_baselines} only represents a possible approximation to the now in-tractable posterior (due to the softmax inverse link function). 
Thus the GP should not be considered as the ground truth nor as the optimal approximation in the classification setting, but is nevertheless useful to give a idea of the level of uncertainty a BNN with a GP prior should provide outside of the support of the data.

\begin{figure*}
    \centering
    \resizebox{\linewidth}{!}{
    \includegraphics[width=\linewidth]{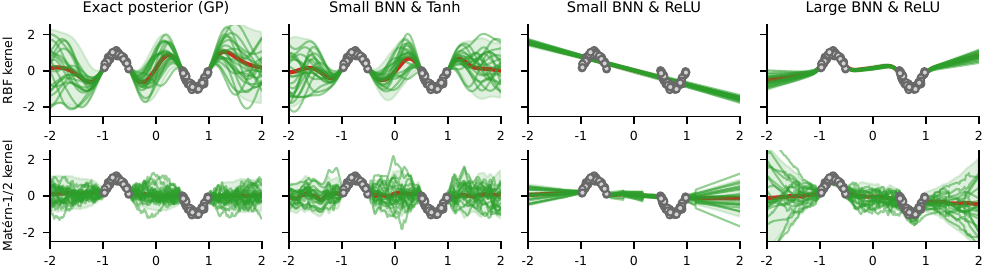}
    }
    \caption{
    Our method requires that the Bayesian neural network (BNN) and Gaussian process (GP) prior share similar inductive biases to provide an accurate approximation to the exact posterior. 
    }
    \label{fig:fsvi_prior_elicitation}
\end{figure*}

\paragraph{Inductive biases}
\cref{fig:fsvi_prior_elicitation} compares GFSVI to the exact posterior across two different priors and three model architectures (details in~\ref{app:sec_synthetic_data}).
We find that the BNN's ability to incorporate the beliefs introduced by the GP prior depends on its size and activation function.
When using piece-wise linear activations (ReLU), small models are prone to underfitting for smooth priors (RBF), and to collapsing uncertainty for rough priors (Matérn-1/2).
By contrast, when using smooth activations (Tanh), smaller models suffice, and they are compatible with most standard GP priors (the results shown in \cref{fig:fsvi_prior_elicitation} extend to RBF, Matérn family, and Rational Quadratic in our experiments).
We also analyzed how the number~$M$ of measurement points affects performance.
\cref{fig:fsvi_varying_n_context_points,fig:kernel_gram_eigendecay} show that capturing the properties of rough GP priors and estimating $D_\text{KL}^\gamma$ with these priors requires larger~$M$.

\subsection{Regression on tabular data}

\begin{table*}[!h]
\scshape
\caption{Test mean square error (MSE) of evaluated methods on regression datasets. We find that GFSVI (ours) also performs competitively in terms of MSE compared to baselines and obtains the best mean rank, matching best the performing methods on nearly all datasets.}
\label{tab:mse}
\resizebox{\linewidth}{!}{
\renewcommand{\arraystretch}{1}
\begin{tabular}{@{}lcccccccccc@{}}
\toprule
\multicolumn{1}{@{}l}{Dataset} & \multicolumn{2}{c}{Function-space priors} & \multicolumn{4}{c}{Weight-space priors} & \multicolumn{3}{c}{Gaussian Processes (Gold Standards)} \\ 
\cmidrule(rl){2-3} \cmidrule(rl){4-7} \cmidrule(rl){8-10}
\multicolumn{1}{@{}l}{} & GFSVI (ours) & \multicolumn{1}{c}{FVI} & TFSVI & MFVI & VIP & Laplace & GWI & Sparse GP & GP  \\ 
\midrule
Boston & \textbf{0.123 $\pm$ 0.021} & \textbf{0.136 $\pm$ 0.022} & 0.995 $\pm$ 0.092 & 0.532 $\pm$ 0.072 & 0.201 $\pm$ 0.056 & 0.203 $\pm$ 0.047 & 0.273 $\pm$ 0.069 & 0.122 $\pm$ 0.014 & 0.115 $\pm$ 0.020 \\
Concrete & \textbf{0.114 $\pm$ 0.008} & \textbf{0.116 $\pm$ 0.004} & 0.389 $\pm$ 0.015 & 0.698 $\pm$ 0.046 & \textbf{0.109 $\pm$ 0.008} & \textbf{0.116 $\pm$ 0.007} & 0.145 $\pm$ 0.017 & 0.399 $\pm$ 0.020 & 0.116 $\pm$ 0.007 \\
Energy & 0.003 $\pm$ 0.000 & 0.003 $\pm$ 0.000 & 0.003 $\pm$ 0.000 & 0.152 $\pm$ 0.024 & 0.043 $\pm$ 0.036 & \textbf{0.002 $\pm$ 0.000} & 0.003 $\pm$ 0.001 & 0.087 $\pm$ 0.005 & 0.087 $\pm$ 0.004 \\
Kin8nm & 0.071 $\pm$ 0.001 & 0.075 $\pm$ 0.003 & 0.073 $\pm$ 0.001 & 0.290 $\pm$ 0.111 & \textbf{0.068 $\pm$ 0.002} & 0.083 $\pm$ 0.001 & 0.071 $\pm$ 0.001 & 0.088 $\pm$ 0.002 & \textit{(infeasible)} \\
Naval & \textbf{0.000 $\pm$ 0.000} & \textbf{0.001 $\pm$ 0.001} & \textbf{0.000 $\pm$ 0.000} & 0.007 $\pm$ 0.003 & 0.002 $\pm$ 0.000 & \textbf{0.000 $\pm$ 0.000} & 0.197 $\pm$ 0.174 & 0.000 $\pm$ 0.000 & \textit{(infeasible)} \\
Power & \textbf{0.052 $\pm$ 0.001} & \textbf{0.054 $\pm$ 0.002} & 0.054 $\pm$ 0.001 & 0.058 $\pm$ 0.002 & \textbf{0.054 $\pm$ 0.002} & \textbf{0.054 $\pm$ 0.002} & 0.052 $\pm$ 0.001 & 0.071 $\pm$ 0.001 & \textit{(infeasible)} \\
Protein & 0.459 $\pm$ 0.005 & 0.466 $\pm$ 0.004 & \textbf{0.429 $\pm$ 0.004} & 0.537 $\pm$ 0.008 & \textbf{0.421 $\pm$ 0.005} & 0.446 $\pm$ 0.006 & 0.425 $\pm$ 0.003 & 0.408 $\pm$ 0.002 & \textit{(infeasible)} \\
Wine & \textbf{0.652 $\pm$ 0.022} & 0.663 $\pm$ 0.009 & 1.297 $\pm$ 0.093 & \textbf{0.655 $\pm$ 0.023} & \textbf{0.627 $\pm$ 0.013} & 0.637 $\pm$ 0.031 & 0.682 $\pm$ 0.048 & 0.607 $\pm$ 0.033 & 0.585 $\pm$ 0.032 \\
Yacht & \textbf{0.003 $\pm$ 0.001} & 0.004 $\pm$ 0.001 & 0.221 $\pm$ 0.037 & 0.682 $\pm$ 0.140 & 0.004 $\pm$ 0.001 & \textbf{0.002 $\pm$ 0.001} & 0.008 $\pm$ 0.003 & 0.399 $\pm$ 0.064 & 0.355 $\pm$ 0.030 \\
Wave & \textbf{0.000 $\pm$ 0.000} & \textbf{0.000 $\pm$ 0.000} & \textbf{0.000 $\pm$ 0.000} & \textbf{0.000 $\pm$ 0.000} & \textbf{0.000 $\pm$ 0.000} & \textbf{0.000 $\pm$ 0.000} & 0.001 $\pm$ 0.001 & 0.000 $\pm$ 0.000 & \textit{(infeasible)} \\
Denmark & \textbf{0.155 $\pm$ 0.004} & 0.287 $\pm$ 0.003 & 0.163 $\pm$ 0.004 & 0.225 $\pm$ 0.003 & 0.189 $\pm$ 0.008 & 0.194 $\pm$ 0.003 & 0.197 $\pm$ 0.004 & 0.260 $\pm$ 0.001 & \textit{(infeasible)} \\
\midrule
Mean rank & 1.364 & 2.000 & 2.182 & 3.182 & 1.636 & 1.727 & - & - & - \\
\bottomrule
\end{tabular}
}
\end{table*}

We present additional regression results reporting the mean square error (MSE) of evaluated methods across the considered baselines, see \cref{tab:mse}.
We find that GFSVI also performs competitively in terms of MSE compared to baselines and obtains the best mean rank, matching best the performing methods on nearly all datasets. 
In particular, we find that using GP priors in the linearized BNN setup with GFSVI yields improvements over the weight-space priors used in TFSVI and that GFSVI performs slightly better than FVI.
Function-space VI methods (TFSVI, GFSVI, FVI) significantly improves over weight-space VI mostly performing similarly to the linearized Laplace approximation. 
Further improvement over baselines are obtained when considering GP priors with GFSVI and FVI.
Finally, GFSVI compares favorably to the GP and sparse GP.

\subsection{Variational measure evaluation}
\label{app:var_measure_eval}

\cref{tab:var_measure_eval} evaluates our inference method by comparing samples drawn from the exact posterior (where computationally feasible) with the approximate posterior obtained with our method (GFSVI).
We follow the setup by \citet{wilson2022evalBNN} and we compute the average per-sample Wasserstein-2 metric samples drawn from a GP posterior with RBF kernel evaluated at the test points, and samples from the approximate posterior of GFSVI, sparse GP and FVI evaluated at the same points and with the same prior. 
More details are provided in \cref{app:sec_measure_eval_details}.
We find that GFSVI approximates the exact posterior more accurately that FVI, obtaining a higher mean rank, but worse than the gold standard sparse GP, which demonstrates to be most accurate.

\begin{table*}[h!]
\scshape
\caption{
Average point-wise Wasserstein-2 distance (lower is better) between exact and approximate posterior of reported methods. GFSVI (ours) provides a more accurate approximation than FVI.
}
\label{tab:var_measure_eval}
\centering
\resizebox{0.8\linewidth}{!}{
\renewcommand{\arraystretch}{1}
\begin{tabular}{@{}lcccccc@{}}
\toprule
Dataset & Boston & Concrete & Energy & Wine & Yacht & Mean rank \\
\midrule
GFSVI (ours) & \textbf{0.0259 $\pm$ 0.0040} & \textbf{0.0499 $\pm$ 0.0029} & \textbf{0.0035 $\pm$ 0.0004} & \textbf{0.0571 $\pm$ 0.0097} & \textbf{0.0036 $\pm$ 0.0006} & 1.0 \\
FVI & 0.0469 $\pm$ 0.0044 & 0.0652 $\pm$ 0.0037 & \textbf{0.0037 $\pm$ 0.0004} & 0.1224 $\pm$ 0.0167 & \textbf{0.0052 $\pm$ 0.0013} & 1.6 \\
\midrule
GP sparse & 0.0074 $\pm$ 0.0022 & 0.0125 $\pm$ 0.0016 & 0.0042 $\pm$ 0.0003 & 0.0170 $\pm$ 0.0035 & 0.0035 $\pm$ 0.0008 & - \\ 
\bottomrule
\end{tabular}
}
\end{table*}

\subsection{Out-of-distribution detection with image data}

\begin{figure*}[!h]
    \centering
    \resizebox{\linewidth}{!}{
    \includegraphics[width=\linewidth]{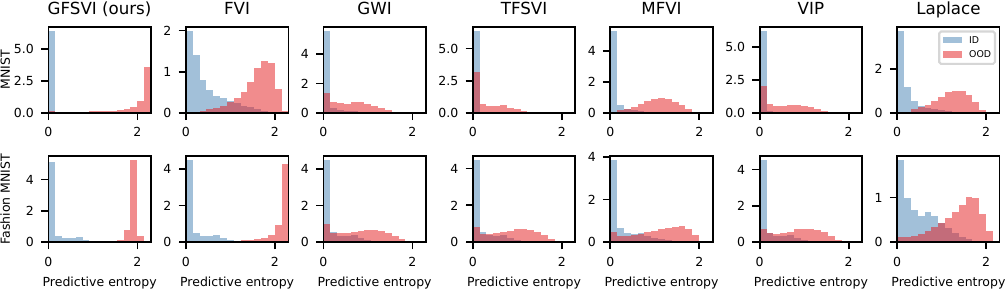}
    }
    \caption{Histograms of the entropy of the mean predictive distribution produced by evaluated methods in the out-of-distribution detection with image data experiment. GFSVI (ours) best partitions in-distribution and out-of-distribution data based on the entropy of its mean predictive distribution.}
    \label{fig:ood_detection_plot}
\end{figure*}

We here show an additional plot from our out-of-distribution detection experiment with image data (details in \ref{app:sec_ood_details}).
\cref{fig:ood_detection_plot} shows the (normalized) histograms of the entropy of the mean prediction produced by each model on the in-distribution (blue) and out-of-distribution (red) data sets considered in our OOD detection experiment.
Methods which estimate the (regularized) KL-divergence in function-space (GFSVI, FVI and TFSVI) use the \textsc{kmnist} measurement distribution. 
We find that the entropy produced by GFSVI on in-distribution data highly peaks around 0 while the entropy produced from out-of-distribution data strongly concentrates around its maximum $\ln (10)$.
GFSVI best partitions ID and OOD data based on predictive entropy improving over the function-space prior (FVI) and weight-space prior (TFSVI, MFVI, Laplace) BNN baselines (see \cref{tab:classification}).

\subsection{Influence of measurement point distribution for image data}
\label{app:influence_rho_image_data}

We present additional results evaluating the influence of the measurement point distribution $\rho$ on the the performance of function-space inference methods when using high-dimensional image data. 
The measurement point distribution are described in \cref{app:sec_classification_details}.
Just like in \citet{rudner2022fsvi}, we find that the choice of measurement point distribution may highly influence the OOD detection accuracy. 
While the expected log-likelihood, accuracy and expected calibration error (ECE) of a model generally remains comparable across measurement point distributions, the OOD accuracy of GFSVI is greatly improved by using samples from \textsc{kmnist} to evaluate the (regularized) KL divergence.
The measurement point distribution determines where the BNN is regularized and thus should be carefully selected especially for high dimensional data.

\begin{table*}[!h]
\scshape
\caption{Influence of the measurement point distribution $\rho$ on expected log-likelihood (log-like.), accuracy (acc.), expected calibration error (ECE) and out-of-distribution detection accuracy (OOD acc.). $\rho$ determines where the BNN will be regularized and strongly influences the out-of-distribution performance of the BNN.}
\label{tab:influence_rho_ood_detection}
\resizebox{\linewidth}{!}{
\renewcommand{\arraystretch}{1}
\begin{tabular}{@{}llcccccccccc@{}}
\toprule
\multirow{1}{*}{\parbox[t]{0pt}{\multirow{2}{*}{\rotatebox[origin=c]{90}{Data\hspace{-0em}}}}}
& Metric & \multicolumn{3}{c}{GFSVI} & \multicolumn{3}{c}{FVI} & \multicolumn{3}{c}{TFSVI}\\
\cmidrule(rl){3-5} \cmidrule(rl){6-8} \cmidrule(rl){9-11}
& & Random & Random Pixel & KMNIST & Random & Random Pixel & KMNIST & Random & Random Pixel & KMNIST \\
\midrule
\multirow{4}{*}{\parbox[t]{0pt}{\multirow{2}{*}{\rotatebox[origin=c]{90}{MNIST\hspace{-1.3em}}}}}
& Log-like.\ ($\uparrow$) & \textbf{-0.033 $\pm$ 0.000} & -0.034 $\pm$ 0.000 & -0.041 $\pm$ 0.000 & -0.145 $\pm$ 0.005 & -0.038 $\pm$ 0.000 & -0.238 $\pm$ 0.006 & -0.047 $\pm$ 0.003 & \textbf{-0.032 $\pm$ 0.001} & -0.041 $\pm$ 0.001 \\
& Acc.\ ($\uparrow$) & \textbf{\hphantom{-}0.992 $\pm$ 0.000} & \hphantom{-}0.989 $\pm$ 0.000 & \hphantom{-}0.991 $\pm$ 0.000 & \hphantom{-}0.976 $\pm$ 0.001 & \hphantom{-}0.988 $\pm$ 0.000 & \hphantom{-}0.943 $\pm$ 0.001 & \hphantom{-}0.989 $\pm$ 0.000 & \hphantom{-}0.989 $\pm$ 0.000 & \hphantom{-}0.989 $\pm$ 0.000 \\
& ECE ($\downarrow$) & \textbf{\hphantom{-}0.002 $\pm$ 0.000} & \hphantom{-}0.004 $\pm$ 0.000 & \hphantom{-}0.006 $\pm$ 0.000 & \hphantom{-}0.064 $\pm$ 0.001 & \hphantom{-}0.003 $\pm$ 0.000 & \hphantom{-}0.073 $\pm$ 0.003 & \hphantom{-}0.007 $\pm$ 0.000 & \hphantom{-}0.003 $\pm$ 0.000 & \hphantom{-}0.006 $\pm$ 0.000 \\
& OOD acc.\ ($\uparrow$) & \hphantom{-}0.921 $\pm$ 0.008 & \hphantom{-}0.868 $\pm$ 0.010 & \textbf{\hphantom{-}0.980 $\pm$ 0.004} & \hphantom{-}0.894 $\pm$ 0.010 & \hphantom{-}0.863 $\pm$ 0.003 & \hphantom{-}0.891 $\pm$ 0.006 & \hphantom{-}0.887 $\pm$ 0.011 & \hphantom{-}0.861 $\pm$ 0.008 & \hphantom{-}0.893 $\pm$ 0.005 \\
\midrule
\multirow{4}{*}{\parbox[t]{0pt}{\multirow{2}{*}{\rotatebox[origin=c]{90}{FMNIST\hspace{-1.5em}}}}}
& Log-like.\ ($\uparrow$) & \textbf{-0.260 $\pm$ 0.003} & \textbf{-0.258 $\pm$ 0.002} & -0.294 $\pm$ 0.006 & -0.300 $\pm$ 0.002 & -0.293 $\pm$ 0.003 & -0.311 $\pm$ 0.005 & -0.261 $\pm$ 0.001 & \textbf{-0.258 $\pm$ 0.001} & -0.261 $\pm$ 0.002 \\
& Acc.\ ($\uparrow$) & \textbf{\hphantom{-}0.910 $\pm$ 0.001} & \hphantom{-}0.908 $\pm$ 0.001 & \textbf{\hphantom{-}0.909 $\pm$ 0.001} & \textbf{\hphantom{-}0.910 $\pm$ 0.002} & \hphantom{-}0.900 $\pm$ 0.001 & \hphantom{-}0.906 $\pm$ 0.002 & \textbf{\hphantom{-}0.909 $\pm$ 0.001} & \hphantom{-}0.908 $\pm$ 0.001 & \hphantom{-}0.907 $\pm$ 0.001 \\
& ECE ($\downarrow$) & \textbf{\hphantom{-}0.020 $\pm$ 0.003} & \hphantom{-}0.022 $\pm$ 0.001 & \hphantom{-}0.042 $\pm$ 0.002 & \hphantom{-}0.027 $\pm$ 0.005 & \textbf{\hphantom{-}0.018 $\pm$ 0.002} & \hphantom{-}0.024 $\pm$ 0.002 & \hphantom{-}0.022 $\pm$ 0.002 & \textbf{\hphantom{-}0.018 $\pm$ 0.001} & \textbf{\hphantom{-}0.021 $\pm$ 0.002} \\
& OOD acc.\ ($\uparrow$) & \hphantom{-}0.853 $\pm$ 0.005 & \hphantom{-}0.867 $\pm$ 0.005 & \textbf{\hphantom{-}0.997 $\pm$ 0.001} & \hphantom{-}0.925 $\pm$ 0.005 & \hphantom{-}0.842 $\pm$ 0.006 & \hphantom{-}0.975 $\pm$ 0.002 & \hphantom{-}0.802 $\pm$ 0.006 & \hphantom{-}0.800 $\pm$ 0.007 & \hphantom{-}0.779 $\pm$ 0.010 \\
\bottomrule
\end{tabular}
}
\end{table*}

\subsection{Input distribution shift with rotated image data}
\label{app:rotated_image_data}

We here provide an experiment evaluating our method's (GFSVI) robustness in detecting input distribution shift. 
We expect the predictive uncertainty of a well-calibrated Bayesian model to be low for in-distribution data and to gradually increase as the input distribution shifts further away from the training data distribution.
To test this property, we follow the setup by \citet{sensoy2018evidential,rudner2022fsvi} and assume like the related work that increasing the rotation angle of images gradually increases the level of input "distribution shift". 
We report the mean and standard-deviation of the average mean predictive entropy of models fit on MNIST \citep{lecun2010mnist} and Fashion MNIST \citep{xiao2017FMNIST} for increasingly large angles of rotation of their respective test data partition.  
We find that GFSVI is confident (low predictive entropy) for images with small rotation angles, and that its predictive entropy increases with the angle.
GFSVI therefore exhibits the expected behavior of a well-calibrated Bayesian model. 
We note that FVI, Laplace and MFVI tend to be under-confident (high predictive entropy) for small rotation angles, which might be a symptom of underfitting further supported by the results in \cref{tab:classification}. 
Also, with the exception of TFSVI, the predictive entropy of baselines across different rotation angles is generally higher than the one produced by GFSVI.
\begin{figure*}[h!]
    \centering
    \resizebox{\linewidth}{!}{
    \includegraphics[width=\linewidth]{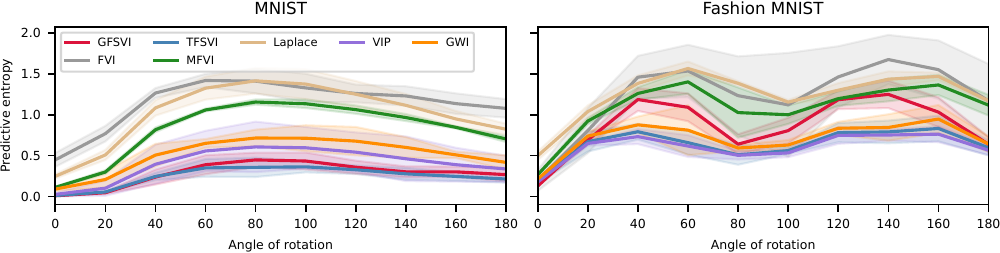}
    }
    \caption{Average predictive entropy of models trained on MNIST and Fashion MNIST and evaluated for different rotation angles of their respective test data partitions. We see that our method (GFSVI) exhibits the behavior of a well-calibrated Bayesian model.}
    \label{fig:input_dist_shift}
\end{figure*}

\subsection{Example of model misspecification with \cite{wild2022gvi}}

\begin{figure*}[h!]
    \centering
    \resizebox{\linewidth}{!}{
    \includegraphics[width=\linewidth,height=1in]{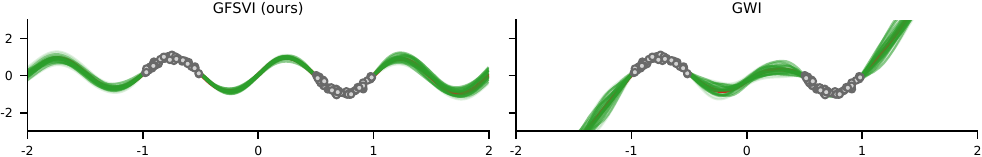}
    }
    \caption{Example of model misspecification when using a periodic GP prior with baseline GWI \citep{wild2022gvi} that does not occur with our method (GFSVI). In GWI, only the posterior covariance is periodic, while the neural network parameterizing the posterior mean results in a function that does not capture the beliefs carried by the (periodic) prior. In contrast, our method accurately captures the GP prior's beliefs and yields a (locally) periodic function.
    }
    \label{fig:gfsvi_vs_gwi_periodic}
\end{figure*}
\cref{fig:gfsvi_vs_gwi_periodic} shows an example of model misspecification when using a periodic GP prior with the baseline GWI \citep{wild2022gvi}. As can be seen in the left panel of the figure, this problem does not occur with our method (GFSVI). While the posterior covariance in GWI reflects the periodicity of the prior, the neural network parametrizing the posterior mean does not result in a periodic function, i.e., the mean does not capture the beliefs specified by the periodic GP prior. In contrast, our method accurately captures the GP prior's beliefs and yields a (locally) periodic function. 

\subsection{Convergence speed on UCI data}
\label{app:convergence_speed}

\begin{figure*}[h]
    \centering
    \resizebox{0.6\linewidth}{!}{
    \includegraphics[width=\linewidth]{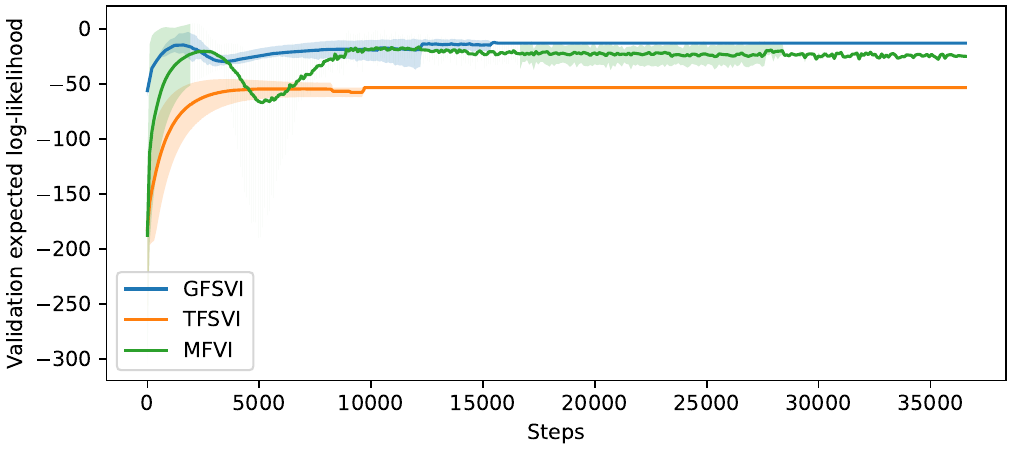}
    }
    \caption{Validation expected log-likelihood of our method (GFSVI) and baselines TFSVI and MFVI. GFSVI (ours) converges on the boston dataset in slightly more steps than TFSVI but in fewer than MFVI.}
    \label{fig:boston_exp_ll_convergence}
\end{figure*}

\begin{wraptable}{r}{7cm}
    \centering
    \scshape
    \caption{Training time of our method GFSVI and baselines MFVI \citep{blundell2015weight} and TFSVI \citep{rudner2022fsvi} on the boston UCI dataset.}
    \resizebox{\linewidth}{!}{
    \begin{tabular}{@{}lccc@{}}
    \toprule
     & GFSVI (ours) & TFSVI & MFVI \\
    \midrule
    Time (s) & 44.15 $\pm$ 1.56 & 36.36 $\pm$ 0.90 & 38.38 $\pm$ 10.80 \\
    \bottomrule
    \end{tabular}
    }
    \label{tab:boston_exp_ll_convergence}
\end{wraptable}
\cref{tab:boston_exp_ll_convergence} shows the training time of our method and baselines MFVI \citep{blundell2015weight} and TFSVI \citep{rudner2022fsvi} on the boston dataset using $M=100$ context points averaged over $5$ cross-validation splits, as well as \cref{fig:boston_exp_ll_convergence} showing the convergence of the validation expected log-likelihood on the boston dataset. Our method converges in more steps than the TFSVI. GFSVI typically takes more time/steps to train that TFSVI as it additionally needs to adapt its features to the beliefs specified by the Gaussian process prior. 

\begin{figure*}[t]
    \centering
    \resizebox{0.6\linewidth}{!}{
    \includegraphics[width=\linewidth]{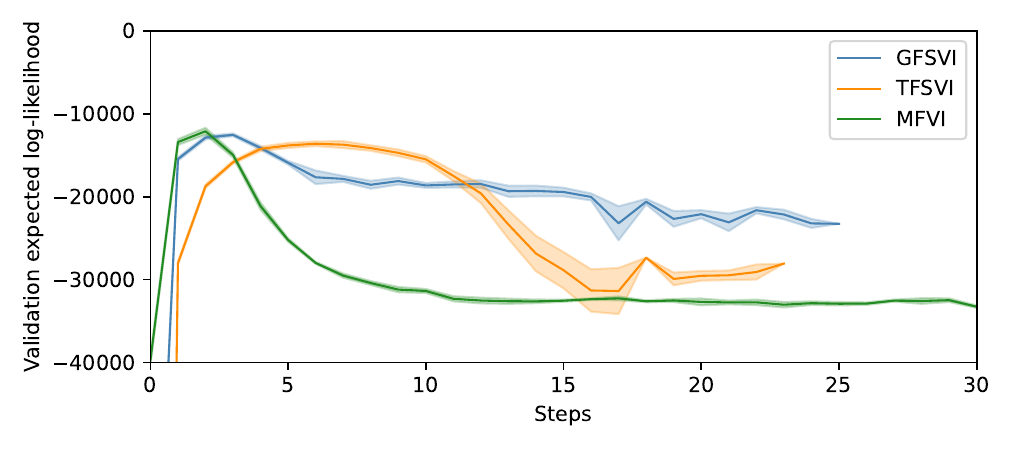}
    }
    \caption{
     Validation expected log-likelihood of our method (GFSVI) and baselines TFSVI and MFVI on MNIST. GFSVI (ours) converges in slightly more steps than TFSVI but in fewer than MFVI.
    }
    \label{fig:mnist_exp_ll_convergence}
\end{figure*}

\subsection{Additional plots for kernel eigenvalue decay}

\begin{figure}[!h]
    \centering
    \begin{tabular}{cc}
    \subfigure[$D=1$]{\includegraphics[width=0.5\linewidth]{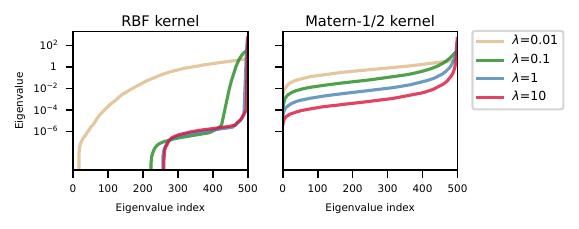}} &
    \subfigure[$D=100$]{\includegraphics[width=0.5\linewidth]{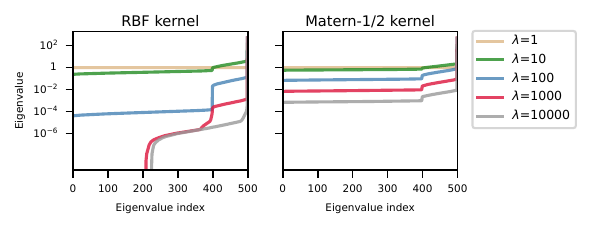}} 
    \end{tabular}
    \caption{Mean eigenvalues of the Gram matrix obtained for different kernels and for varying length-scales over $10$ draws from a uniform distribution on $[-2, 2 ]^D$. The mean eigenvalues are arranged in increasing order. The eigenvalues of the Gram matrix associated with the smooth RBF kernel decays much faster than those of the Matérn-1/2. Furthermore, the eigenvalues decay at a slower rate in high dimensions (D=100).}
    \label{fig:kernel_gram_eigendecay}
\end{figure}

\cref{fig:kernel_gram_eigendecay} shows a plot demonstrating the decay rate of the eigenvalues of RBF and Matérn-1/2 kernels evaluated at points sampled uniformly over $\mathcal{X}$ . 
The rate of decay of covariance operator's eigenvalues gives important information about the smoothness of stationary kernels \citep{williams2006gaussian} and that increased smoothness of the kernel leads to faster decay of eigenvalues \citet{Santin2016approxEigenfunc}.
For instance, RBF covariance operator eigenvalues decay at near exponential rate independent of the underlying measure \citep{belkin2018approximation} and Matérn kernels eigenvalues decay polynomialy \citep{chen2021gaussian}.
We find that the kernel evaluated at points sampled from a uniform distribution over $\mathcal{X}$ share this same behavior (see \cref{fig:kernel_gram_eigendecay}).

\subsection{Additional plots for choosing $\gamma$ in $D_\text{\normalfont KL}^\gamma$}
\label{seq:choosing_gamma}

\begin{figure*}[h!]
    \centering
    \resizebox{\linewidth}{!}{
    \includegraphics[width=\linewidth]{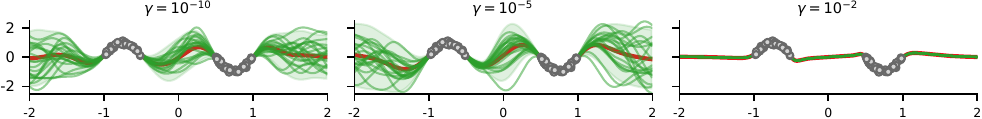}
    }
    \caption{The $\gamma$ parameter of the regularized KL divergence controls the magnitude of the regularizer in the objective and should be small enough to provide strong regularization.}
    \label{fig:fsvi_influence_gamma}
\end{figure*}

\begin{figure}[h!]
\centering
\noindent
\begin{minipage}[t]{0.45\linewidth}
\includegraphics[width=\linewidth]{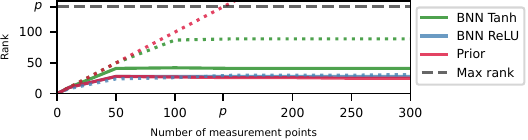}
\captionof{figure}{The BNN's covariance adaptation to the prior's covariance rank depends on its activation function. BNNs fit with a RBF prior (full) show lower rank than with a Matérn-1/2 (dotted).}
\label{fig:approx_posterior_rank}
\end{minipage}%
\hfill
\begin{minipage}[t]{0.53\linewidth}
\includegraphics[width=\linewidth]{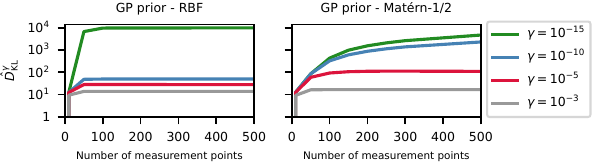}
\captionof{figure}{$\gamma$ explicitly controls the magnitude of the regularized KL-divergence $D_\text{KL}^\gamma$. Rougher priors (Matérn-1/2) require more measurement points to accurately estimate $D_\text{KL}^\gamma$ than smooth priors (RBF).}
\label{fig:regkl_vs_gamma}
\end{minipage}%
\end{figure}

The $\gamma$ parameter controls the magnitude of the regularized KL divergence (see \cref{fig:regkl_vs_gamma}) and adjusts the relative weight of the regularized KL divergence and expected log-likelihood term in the training objective (see \cref{fig:fsvi_influence_gamma}).  
Furthermore, $\gamma$ also acts as "jitter" preventing numerical errors. 
We recommend choosing $\gamma$ large enough to avoid numerical errors while remaining small enough to provide strong regularization. 

\end{document}